\documentclass[11pt]{article}

\usepackage[preprint]{acl}

\usepackage{times}
\usepackage{latexsym}

\usepackage[T1]{fontenc}

\usepackage[utf8]{inputenc}

\usepackage{microtype}

\usepackage{inconsolata}

\usepackage{graphicx}
\usepackage{amsmath}
\usepackage{amssymb}
\usepackage{amsthm}
\usepackage{algorithm}
\usepackage{algorithmic}
\newtheorem{definition}{Definition}
\newtheorem{proposition}{Proposition}
\usepackage{multirow}
\usepackage{arydshln}   
\usepackage[table]{xcolor} 
\usepackage{booktabs}   
\usepackage{graphicx}
\usepackage{caption}
\usepackage{subcaption}
\usepackage{pifont}
\usepackage[strings]{underscore}

\newcommand{\cmark}{\ding{51}}
\newcommand{\xmark}{\ding{55}}
\usepackage{siunitx}
\usepackage{dblfloatfix}
\usepackage[most]{tcolorbox}
\title{LDDR: Linear-DPP Dynamic Resolution Frame Sampling for Video MLLMs}

\author{%
  \textbf{Jingfeng Chen$^{1}$ \quad
  Jiawen Qian$^{2}$ \quad
  Wendi Deng$^{2}$ \quad
  Yinuo Guo$^{3}$} \\
  \textbf{Jiaqi Yu$^{2}$ \quad
  Sicong Leng$^{4}$ \quad
  Raghuveer Thirukovalluru$^{5}$ \quad
  Bhuwan Dhingra$^{5}$} \\
  \\
  $^{1}$Carnegie Mellon University 
  $^{2}$Individual Researcher
  $^{3}$National University Singapore
  \\
  $^{4}$Nanyang Technological University 
  $^{5}$Duke University
}

\begin{document}
\maketitle
\begin{abstract}
Multimodal large language models, when used for long video understanding, require selecting an informative frame set from massive and redundant candidates. Limited by the model's visual-token budget and reasoning ability, frame selection is therefore essential to reduce input size while preserving the most relevant information for accurate video reasoning. Existing frame-sampling methods like uniform sampling or relevance/diversity-sampling often miss important events or retain redundant frames. On the other hand, agentic frame selection methods incur very high computational costs. Further, these methods allocate equal tokens to all selected frames, regardless of their importance. We therefore propose \textbf{LDDR} (\textbf{L}inear \textbf{D}PP \textbf{D}ynamic \textbf{R}esolution), which is a training-free and budget-aware video frame sampling framework. It is computationally efficient and allows non-uniform token budget allocation within the candidate frame set. Specifically, LDDR performs query-aware Determinantal Point Process (DPP) frame selection in a task-conditioned feature space, achieving a 3$\times$ runtime speedup over standard DPP baselines. It further introduces a Group DPP importance score to guide frame retention and dynamic resolution allocation: assigning more tokens to informative and non-redundant frames while downscaling or pruning less useful ones. We test LDDR on four popular video benchmarks spanning short, medium, and long-range videos. LDDR achieves a 2.5-point gain in low-budget settings and a 1.6-point gain in high-budget scenarios. These improvements are consistently observed across multiple MLLM families and sizes. Further manual qualitative analyses are included to prove the method's effectiveness.

\end{abstract}

\section{Introduction}
    Using Multimodal Large Language Models (MLLMs) for video understanding requires compressing thousands of frames into a much smaller set of informative visual tokens. Although recent MLLM backbones can support increasingly large visual input, prior work has shown simply adding more frames does not necessarily improve video reasoning ability~\citep{sun2025mdp3}. This is because effective video reasoning depends on selecting the most relevant and informative frames, as long videos often contain substantial redundancy: many frames provide little new information, increase computational cost, and distract the model from critical evidence~\citep{yu2024frame}. Therefore, frame sampling plays a key role in video processing. It determines which visual evidence is preserved and how effectively the MLLMs can reason over the video. Despite its significance, recent MLLM backbones~\citep{bai2025qwen25vltechnicalreport, bai2025qwen3, wang2025internvl3} by default rely on a query-agnostic uniform sampling of video frames.
    
    Uniform sampling overlooks that only a subset of frames is relevant to a given query, leading to redundancy and the risk of missing critical evidence. This has driven a shift to query-driven frame selection, where sampling is conditioned on the input query. Recent approaches have explored Top-$K$ query-frame relevance sampling~\citep{yu2023self, liang2024keyvideollm, wang2024weakly}, by using CLIP-style text-visual encoders \citep{radford2021learning, zhai2023sigmoid, zhang2024long} to independently score and select frames based on their relevance to the user input. Such point-wise selection methods fail to account for inter-frame dependencies, often leading to redundancy and incomplete interpretation of the video flow. To mitigate this, later work has proposed agentic exploration~\citep{yang2412vca, panzoomv, zuo2025videolucy, tang2026tspo} or learned frame selectors~\citep{hu2025m} for temporally coherent and less redundant frame selection. However, these methods introduce huge training and inference overhead.

    Among training-free approaches, Determinantal Point Processes (DPPs)~\citep{macchi1975coincidence,gong2014diverse} have proven as one of the most effective frameworks for selecting query-relevant frames while also minimizing redundancy. However, existing DPP-based approaches, such as MDP3~\citep{sun2025mdp3} exhibit two key limitations. First, MDP3 performs DPP inference in kernel space, where kernel construction and updates scale quadratically with the number of video frames, making global frame selection expensive for long videos. To mitigate this, MDP3 performs DPP-based selection over temporally chunked video segments. Secondly, MDP3 allocates the uniform token budget to all selected frames, despite large differences in their visual-information complexity. With modern vision encoders supporting dynamic resolutions~\citep{bai2025qwen3,an2025llava}, video understanding can instead be formulated as a joint optimization over frame selection and per-frame token allocation.

    We therefore propose \textbf{LDDR} (\textbf{L}inear \textbf{D}PP \textbf{D}ynamic \textbf{R}esolution), a training-free framework for budget-aware long-video understanding. LDDR performs DPP-based selection directly in frame-feature space rather than kernel space, reducing complexity from quadratic to linear in the number of video frames. LDDR further proposes a \textbf{Group DPP (GD)} importance score to measure the marginal contribution of each selected frame and guide adaptive per-frame token allocation. Highly relevant and non-redundant frames receive larger token budgets, while redundant or less informative frames are downscaled or pruned. Our contributions are summarized as follows:

    \begin{itemize}
    \item \textbf{Linear DPP Selection.} We perform query-aware DPP frame selection jointly over all video frames, avoiding temporal chunking and explicit kernel construction, and enabling frame selection with linear complexity in the number of original video frames.

    \item \textbf{Group DPP Importance for Dynamic Resolution.} We propose a Group DPP importance score based on each frame's marginal determinant contribution and use it to jointly guide frame retention and adaptive per-frame token allocation.

    \item \textbf{Comprehensive Empirical Validation.} We evaluate LDDR across four long-video benchmarks: Video-MME, LongVideoBench, LVBench, and MLVU; and multiple MLLM backbones. LDDR achieves average gains of +2.5 points under budget-constrained settings and +1.6 points in high-budget settings.
    
    \item \textbf{Efficient and Scalable.} Despite achieving strong performance gains, LDDR remains highly efficient and scalable, substantially outperforming existing baselines in runtime. 
    
\end{itemize}

\section{Related Work}
\textbf{Multimodal Video Understanding.}
Multimodal large language models (MLLMs) have demonstrated remarkable capabilities. Generally, existing strategies to handle long-video input can be categorized into three streams: video token compression after extensive frame sampling~\citep{song2024moviechat, jin2024chat, li2025less, li2024llama, shao2025holitom, zhang2025flexselect, shi2025slow, luoenhancing}, heuristic or learned frame selection~\citep{hu2025m, tang2025adaptive, tang2026tspo, zhu2025focus, gong2014diverse, sun2025mdp3}, and agentic frameworks that employ iterative exploration, feedback, memory-update loops to retrieve informative frames along with query reasoning~\citep{wang2024videoagent, wang2025videotree, yang2412vca, panzoomv, zuo2025videolucy, luo2024video}.

\textbf{Frame Selection and Redundancy Reduction.} While agentic frameworks can achieve high accuracy through iterative reasoning, they often incur substantial computational overhead. While token pruning majorly reduces the number of tokens that frames transform into. Regarding frame selection, traditional methods typically rely on Top-$K$ query–frame relevance sampling~\citep{yu2023self, wang2024weakly, liang2024keyvideollm}. Subsequent work jointly considers relevance and temporal coverage. For instance, AKS~\cite{tang2025adaptive} introduces top relevance under binary-divided sections, while FOCUS~\cite{zhu2025focus} employs a two-stage bandit selection mechanism. Furthermore, chunked DPP-based approaches such as DSS~\cite{gong2014diverse} and MDP3~\cite{sun2025mdp3} are widely applied to select diverse subsets of frames. Recent advances in image token pruning \cite{zhang2025beyond} suggest that globally optimized DPPs can yield superior performance, but these methods are not efficient for long-video settings due to quadratic computational complexity.

\begin{figure*}[t] \centering \includegraphics[width=0.99\textwidth]{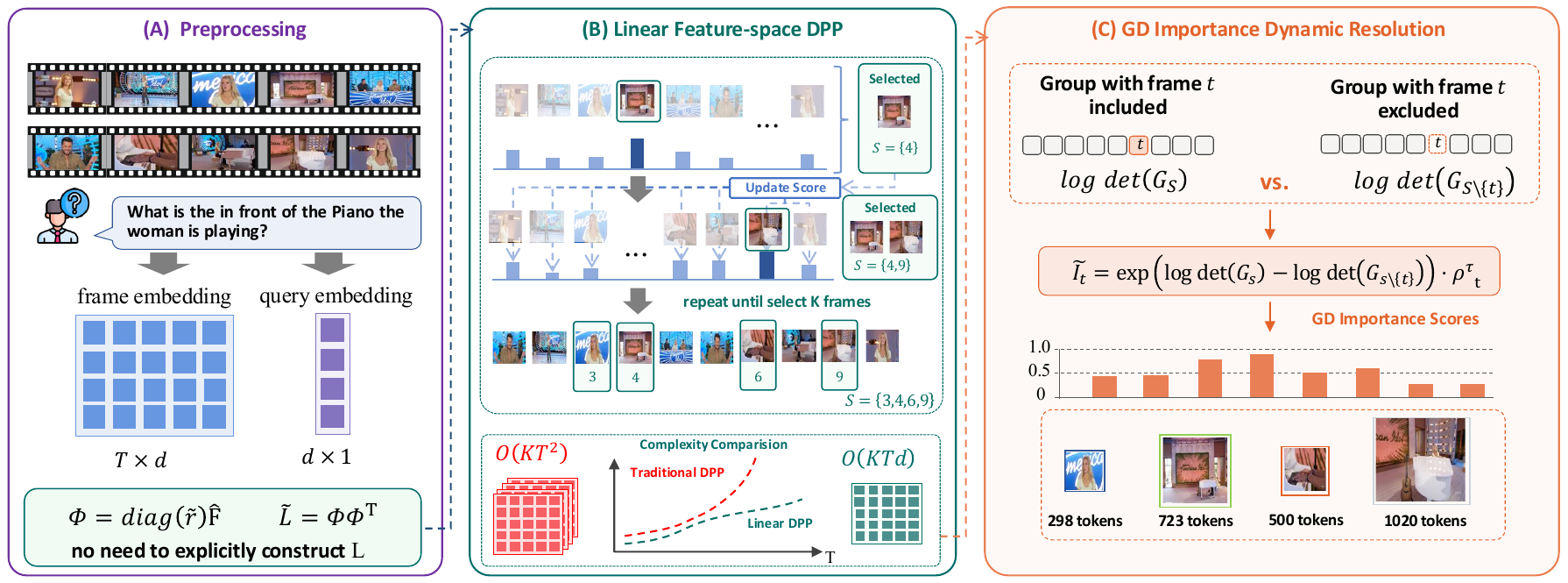} \caption{LDDR Overview. LDDR first extracts frame and query embeddings, then applies Linear Feature-space DPP to select a diverse and query-relevant candidate frame set. GD-importance-based dynamic resolution further allocates visual tokens adaptively across selected frames.} \label{fig:my_image} \end{figure*}

\textbf{Dynamic Frame Resolution.}
Dynamic resolution across frames allows balancing efficiency with visual detail for video understanding.
Prior works, such as Q-frame \citep{zhang2025q}, Anchor Frames \citep{sun2025from}, and ResAdapt \citep{liao2026resadapt}, allocate variable visual tokens
across frames via relevance-based scoring or reinforcement learning.
These methods either require additional training or modification of the model structure.
Recent models like Qwen2.5-VL \citep{bai2025qwen25vltechnicalreport} and LLaVA-OneVision-1.5 \citep{an2025llava} support flexible token allocation from variable-resolution inputs, but do not explicitly address how to allocate tokens under a global budget.
Therefore, we develop a training-free, budget-aware pipeline that couples global diversity-preserving frame selection with GD-guided token allocation,
relying on the same DPP feature space to guide both.

\section{Methodology}
\subsection{LDDR Overview}
Figure~\ref{fig:my_image} shows our overall pipeline. We first utilize a Linear DPP to construct a query-aware and diverse candidate set of frames $\mathcal S$, and then optimize their resolutions under a total visual-token budget $C_{\mathrm{total}}$. Instead of selecting a fixed number of frames, we treat the per-frame visual-token count $w_t$ as the control variable. Each candidate is assigned tokens within a range $[w_{\min}, w_{\max}]$, where $w_{\min}$ ensures a valid visual resolution for effective MLLM processing and $w_{\max}$ limits the computational cost. Tokens are allocated based on DPP in-group contribution, such that more informative candidates receive higher resolution.
The rest of this section is organized as follows:
Section \ref{ss:background} reviews DPPs and their application to frame selection; and
Section \ref{ss:implementation} presents our linear DPP formulation for video frame selection (\ref{sss:lineardpp}) and dynamic resolution allocation (\ref{sss:dynamic_res}) pipeline.

\subsection{DPP Background and Problem Setup}\label{ss:background}
Determinantal Point Processes (DPPs), originally used
to characterize fermion repulsion in quantum mechanics~\cite{macchi1975coincidence},
have been widely used for subset selection, including recommendation,
summarization~\cite{chen2018fast, celis2018fair}, video frame sampling~\cite{sun2025mdp3,gong2014diverse},
and vision token pruning~\cite{zhang2025beyond}. DPPs provide a probabilistic framework for subset selection that balances \textit{quality} and \textit{diversity}. Given a ground set of items $\mathcal{Y} = \{1, \dots, T\}$, a DPP defines a probability distribution over subsets $S \subseteq \mathcal{Y}$ as:
\begin{equation}
P(S) \propto \det(L_S)
\end{equation}
where $L \in \mathbb{R}^{T \times T}$ is a positive semidefinite kernel matrix that captures how two items in the set interact, and $L_S$ denotes the principal submatrix indexed by $S$. The determinant has a geometric interpretation: it measures the volume spanned by the feature vectors of the selected items. Therefore, subsets containing similar (redundant) items have a lower probability, naturally promoting diversity.

We follow  \citet{zhang2025beyond} to construct a query-aware DPP kernel $\mathbf{L} \in \mathbb{R}^{T \times T}$ to encode both frame relevance and inter-frame diversity, where $T$ denotes the number of video frames.
Frame embeddings $\mathbf{H} \in \mathbb{R}^{T \times d}$ and the query embedding $\mathbf{q} \in \mathbb{R}^d$ are obtained using CLIP-style vision and text encoders in a shared embedding space.
Let $\hat{\mathbf{F}}$ denote the $\ell_2$-normalized
frame embeddings,
and 
$\mathbf{S}=\hat{\mathbf{F}}\hat{\mathbf{F}}^\top \in \mathbb{R}^{T \times T}$ be the pairwise-frame
similarity.
Then, query relevance is computed via cosine similarity between each frame embedding and the query embedding: $r_t=\frac{\mathbf{h_t}^{\top}\mathbf{q}}{\|\mathbf{h_t}\|\|\mathbf{q}\|}$, followed by MinMax normalization to obtain $\tilde{\mathbf{r}} \in \mathbb{R}^T$.
The query-aware kernel is then defined as $\mathbf{L}=\mathrm{diag}(\tilde{\mathbf{r}})\,\mathbf{S}\,\mathrm{diag}(\tilde{\mathbf{r}})$, where $\mathbf{S}$ captures redundancy and $\tilde{\mathbf{r}}$ encodes query-dependent importance.
The most relevant and diverse subset is finally obtained through greedy MAP inference:
\begin{equation}
S^* = \text{argmax}_{S \subseteq \mathcal{Y}, \, |S|= K} \log \det(L_S),
\end{equation}
where $L_S$ is the principal submatrix indexed by $S$, and $K$ is the frame budget.
Standard DPP solvers use a greedy algorithm in kernel space \citep{chen2018fast, zhang2025beyond} which iteratively selects the item with the largest marginal gain in the log-determinant above. This requires $\mathcal{O}(T^2)$ space and $\mathcal{O}(T^2 d + TK^2)$ time. However, the quadratic complexity in terms of $T$ makes long video processing time inefficient. This motivates the efficient LDDR design that we introduce next.

\subsection{LDDR Technical Details}\label{ss:implementation}
\subsubsection{Linear DPP Selection for Frame Candidate Set}\label{sss:lineardpp}
Prior work has explored several strategies to make DPP solutions more efficient. Those including low-rank approximations \citep{gartrell2017low}, incremental updates \citep{chen2018fast}, lazy Cholesky methods \citep{hemmi2022lazy}, and chunking \citep{gong2014diverse,sun2025mdp3}
But they are still designed to accelerate inference over an item-to-item kernel. Therefore, the full kernel update is still required during computation. The determinant-volume interpretation of
DPPs~\citep{Kulesza_2012} provides the geometric
foundation for our formulation.
We designed a feature-space implementation
of global query-aware video frame selection that directly maintains an orthonormal basis of selected frame features and then updates candidate residuals. This approach eliminates explicit frame-frame kernel construction, and the computational cost scales linearly with video length. Specifically, we design the frame kernel as a Gram matrix of query-weighted frame features:

\begin{equation}
\begin{aligned}
\mathbf L
&=
\mathrm{diag}(\tilde{\mathbf r})
\hat{\mathbf F}\hat{\mathbf F}^{\top}
\mathrm{diag}(\tilde{\mathbf r})
=
\mathbf{\Phi}\mathbf{\Phi}^{\top},\\
\mathbf{\Phi}
&=
\mathrm{diag}(\tilde{\mathbf r})\hat{\mathbf F},\\
{\Longrightarrow}
\qquad
L_{st}
&=
\mathbf{\Phi}_{s,:}\mathbf{\Phi}_{t,:}^{\top}
\end{aligned}
\end{equation}

We perform greedy MAP selection directly in the feature space $\mathbf{\Phi}$, rather than building a dense $T\times T$ kernel. For the kernel $\mathbf L=\mathbf{\Phi}\mathbf{\Phi}^{\top}$, this feature-space procedure produces the same greedy selection sequence as kernel-space DPP selection, while reducing the cost to $\mathcal{O}(KTd)$ time and $\mathcal{O}(Td)$ memory. The algorithm maintains an orthonormal basis $\mathbf C \in \mathbb{R}^{K \times d}$ for the selected directions and residual gains $\mathbf d \in \mathbb{R}^T$ for the remaining frames. 

In each iteration, an unselected frame with the largest residual gain is added to the set, and the gains of the remaining frames are updated by removing their projection onto the newly selected direction. This update penalizes frames that are redundant with the selected set: later iterations favor frames that are both query-relevant and diverse. A proof of equivalence to greedy kernel-space DPP selection is provided in Appendix~\ref{appendix:equivalence}. 

\begin{algorithm}[H]
\caption{Feature-space Greedy DPP Inference}
\begin{algorithmic}[1]
\STATE \textbf{Input:} Feature matrix $\mathbf{\Phi} \in \mathbb{R}^{T \times d}$, budget $K$
\STATE \textbf{Initialize:} $d_t \leftarrow \|\mathbf{\Phi}_{t,:}\|_2^2$ for all $t$, $\mathcal S \leftarrow \emptyset$, $\mathbf C \leftarrow [\,]$
\FOR{$i = 1$ to $K$}
    \STATE $j \leftarrow \arg\max_{t \notin \mathcal S} d_t$
    \STATE $\mathcal S \leftarrow \mathcal S \cup \{j\}$
    \STATE $\mathbf v \leftarrow \mathbf{\Phi}_{j,:} - \sum_{k=1}^{i-1} (\mathbf c_k \mathbf{\Phi}_{j,:}^{\top}) \mathbf c_k$
    \STATE $\mathbf c_i \leftarrow \mathbf v / \sqrt{d_j}$
    \FOR{each $t \notin \mathcal S$}
        \STATE $d_t \leftarrow d_t - (\mathbf{\Phi}_{t,:} \mathbf c_i^{\top})^2$
    \ENDFOR
\ENDFOR
\STATE \textbf{Output:} Selected Candidate set $\mathcal S$
\end{algorithmic}
\end{algorithm}

The residual gain during iteration reflects the incremental contribution at selection time; it is order-dependent and does not fairly quantify the frame value. We therefore propose a GD importance score.

\subsubsection{GD Importance Guided Dynamic Resolution}\label{sss:dynamic_res}

From the DPP-selected candidate set $\mathcal S$ of size $K$, our goal is to decide both how many frames to retain and how many visual tokens to assign to each retained frame within the total budget $C_{\mathrm{total}}$. The value $k^*$ denotes the final number of retained frames, and $\mathcal S^*$ denotes the final retained frame set. We propose a
\textbf{density-aware GD importance} to quantify the token assignment.
\begin{definition}[GD Importance]
For a selected set $\mathcal S$, the Group DPP (GD) importance of a frame $t\in\mathcal S$ is defined as its marginal determinant contribution:

\begin{equation}
\label{eq:GD}
\begin{aligned}
\mathcal I_t
&=
\exp\!\left(
\log\det(\mathbf G_{\mathcal S})
-
\log\det(\mathbf G_{\mathcal S\setminus\{t\}})
\right),\\
\mathbf G_{\mathcal S}
&=
\mathbf\Phi_{\mathcal S}
\mathbf\Phi_{\mathcal S}^{\top}.
\end{aligned}
\end{equation}
\end{definition}

\begin{proposition}[Residual form]
\label{prop:gd_residual}
For any $t\in\mathcal S$,
\begin{equation}
\begin{aligned}
\mathcal I_t
&=
\left\|
\left(
\mathbf I
-
\mathbf P_{\mathcal S\setminus\{t\}}
\right)
\boldsymbol\phi_t
\right\|_2^2.
\end{aligned}
\end{equation}
Here, $\mathbf P_{\mathcal S\setminus\{t\}}$ projects onto the span of
$\{\boldsymbol\phi_j : j\in\mathcal S\setminus\{t\}\}$.
\end{proposition}

Proposition~\ref{prop:gd_residual} shows that $\mathcal I_t$ measures the
residual uniqueness of frame $t$; its proof and geometric interpretation are
provided in Appendix~\ref{appendix:gd_residual_uniqueness}.
In practice, we also apply a density prior to the GD Score
$\tilde{\mathcal I}_t=\mathcal I_t\rho_t^\tau$, where 
$\rho_t=\frac{\|\boldsymbol\phi_t\|_2^2}{|\mathcal S|^{-1}\sum_{j\in\mathcal S}\|\boldsymbol\phi_j\|_2^2}$ and $\tau$ controls the prior strength. In all experiments, $\tau=1.0$ is used without tuning on datasets, and we report a sensitivity analysis in Appendix~\ref{density-prior}.

Specifically, we first sort candidate frames in descending order of their density-aware GD importance scores $\tilde{\mathcal I}_t$. Let $\mathcal S_k$ denote the top-$k$ prefix in this sorted order. To determine how many frames can be retained, we perform a binary search over $k$. For each candidate prefix $\mathcal S_k$, we normalize the importance scores within the prefix as $\alpha_t^{(k)}=\tilde{\mathcal I}_t/\sum_{j\in\mathcal S_k}\tilde{\mathcal I}_j$ and compute the clamped token allocation $\hat w_t^{(k)}=\min\{w_{\max},\max\{w_{\min},C_{\mathrm{total}}\alpha_t^{(k)}\}\}$, ensuring that every retained frame has least $w_{\min}$ tokens. The prefix is considered feasible if $\sum_{t\in\mathcal S_k}\hat w_t^{(k)}\le C_{\mathrm{total}}$. We then select the largest feasible prefix size $k^*=\max\{k:\sum_{t\in\mathcal S_k}\hat w_t^{(k)}\le C_{\mathrm{total}}\}$ and retain $\mathcal S^*=\mathcal S_{k^*}$, while candidates outside this prefix are evicted. This procedure ensures that the retained frames are the most important ones under the GD scores, and that their token allocations remain within the valid range $[w_{\min},w_{\max}]$.

\section{Experiments}\label{sec:experiments}

\textbf{Models}: We evaluate our LDDR on a diverse set of MLLMs, including Qwen2.5-VL-7B~\cite{bai2025qwen25vltechnicalreport}, Qwen3-VL-8B~\cite{bai2025qwen3}, LLaVA-OneVision-1.5-8B~\cite{an2025llava}, and InternVL3.5 (4B/8B)~\cite{wang2025internvl3}. These models differ in visual encoders and reasoning capabilities. Our primary method is LDDR. For models that support dynamic resolution, including Qwen-VL and LLaVA variants, we apply the full LDDR framework. For models without dynamic resolution support, we use the fixed-resolution variant, LD. We evaluate all models under unified inference settings without additional training, following official \texttt{lmms-eval}~\cite{zhang2407lmms}.

\textbf{Benchmarks}: We conduct experiments on four popular long-video benchmarks: Video-MME~\cite{fu2025video}, LongVideoBench~\cite{wu2024longvideobench}, LVBench~\cite{wang2025lvbench}, and MLVU~\cite{zhou2025mlvu}. 

\textbf{Baselines}: We compare our LDDR with the latest frame selection methods, including uniform sampling, query-aware dynamic resolution: Q-frame~\citep{zhang2025q}, binary partition relevance sampling: AKS~\cite{tang2025adaptive}, optimal arm selection: FOCUS~\cite{zhu2025focus}, and chunked DPP with dynamic programming: MDP3~\cite{sun2025mdp3}. As all of these methods will require a text-image encoder and their original implementations use different CLIP variants, we standardize the pre-encoding with LongCLIP to guarantee fairness. This standardization does not disadvantage the baselines; in fact, we observe that the LongCLIP-based results are generally better than their originally reported results. More details are provided in Appendix~\ref{appendix:implementation}.

\begin{table*}[!t]
\centering
\caption{Main results across four benchmarks and five MLLM backbones under matched visual-token budgets. * denotes models supporting dynamic resolution; $\dagger$ denotes LD (linear DPP frame selection), and $\ddagger$ denotes LDDR (GD-guided dynamic resolution). LD improves over baselines, and LDDR achieves further gains in 12/16 cases.}
\label{tab:main_results_full}
\providecommand{\LDDRbase}[1]{{\scriptsize #1}}
\providecommand{\LDDRgain}[1]{\textsuperscript{\tiny$\uparrow$#1}}
\footnotesize
\setlength{\tabcolsep}{2.9pt}
\renewcommand{\arraystretch}{0.6}
\begin{tabular}{cc l llll lllll l l}
\toprule
\multirow{2}{*}{\textbf{Model}} & \multirow{2}{*}{\textbf{\#F}} & \textbf{Method}
& \multicolumn{4}{c}{\textbf{Video-MME}}
& \multicolumn{5}{c}{\textbf{LongVideoBench}}
& \multirow{2}{*}{\textbf{LVBench}} & \multirow{2}{*}{\textbf{MLVU}} \\
\cmidrule(lr){4-7}\cmidrule(lr){8-12}
& & \hfill / \textit{subset}
& \textit{Short} & \textit{Med} & \textit{Long} & \textit{Overall}
& \textit{15s} & \textit{60s} & \textit{600s} & \textit{3600s} & \textit{Overall} & & \\
\midrule
 & \multirow{7}{*}{8} & \LDDRbase{Uniform} & \LDDRbase{65.1} & \LDDRbase{51.1} & \LDDRbase{48.1} & \LDDRbase{54.78} & \LDDRbase{68.3} & \LDDRbase{66.3} & \LDDRbase{51.2} & \LDDRbase{46.8} & \LDDRbase{53.70} & \LDDRbase{33.89} & \LDDRbase{53.31} \\
 &  & \LDDRbase{AKS} & \LDDRbase{70.4} & \LDDRbase{57.1} & \LDDRbase{49.2} & \LDDRbase{58.93} & \LDDRbase{66.7} & \LDDRbase{70.4} & \LDDRbase{59.5} & \LDDRbase{52.7} & \LDDRbase{59.01} & \LDDRbase{40.54} & \LDDRbase{65.54} \\
 &  & \LDDRbase{Q-frame} & \LDDRbase{69.8} & \LDDRbase{60.7} & \LDDRbase{51.1} & \LDDRbase{59.07} & \LDDRbase{70.9} & \LDDRbase{68.6} & \LDDRbase{55.1} & \LDDRbase{49.1} & \LDDRbase{56.54} & \LDDRbase{36.15} & \LDDRbase{58.49} \\
 &  & \LDDRbase{FOCUS} & \LDDRbase{69.0} & \LDDRbase{54.1} & \LDDRbase{43.7} & \LDDRbase{55.59} & \LDDRbase{58.7} & \LDDRbase{69.6} & \LDDRbase{59.0} & \LDDRbase{50.4} & \LDDRbase{56.54} & \LDDRbase{37.18} & \LDDRbase{63.01} \\
 &  & \LDDRbase{MDP3} & \LDDRbase{71.9} & \LDDRbase{55.1} & \LDDRbase{49.8} & \LDDRbase{59.88} & \LDDRbase{66.7} & \LDDRbase{72.7} & \LDDRbase{59.2} & \LDDRbase{50.9} & \LDDRbase{58.49} & \LDDRbase{40.28} & \LDDRbase{66.21} \\
\noalign{\vskip 2pt}\cdashline{3-14}[5pt/1.2pt]\noalign{\vskip 2pt}
\rowcolor{gray!10}
 &  & \LDDRbase{$\dagger$ \textbf{LD}} & \LDDRbase{73.3} & \LDDRbase{57.0} & \LDDRbase{49.6} & \LDDRbase{59.96\LDDRgain{5.2}} & \LDDRbase{66.7} & \LDDRbase{69.8} & \LDDRbase{59.2} & \LDDRbase{54.8} & \LDDRbase{59.76\LDDRgain{6.1}} & \LDDRbase{43.83\LDDRgain{9.9}} & \LDDRbase{\textbf{66.63}\LDDRgain{13.3}} \\
\rowcolor{gray!10}
 &  & \LDDRbase{$\ddagger$ \textbf{LDDR}} & \LDDRbase{72.9} & \LDDRbase{59.8} & \LDDRbase{52.8} & \LDDRbase{\textbf{61.82}\LDDRgain{7.0}} & \LDDRbase{69.3} & \LDDRbase{73.3} & \LDDRbase{60.7} & \LDDRbase{55.9} & \LDDRbase{\textbf{61.48}\LDDRgain{7.8}} & \LDDRbase{\textbf{45.51}\LDDRgain{11.6}} & \LDDRbase{66.04\LDDRgain{12.7}} \\
\cmidrule(lr){2-14}
 & \multirow{7}{*}{32} & \LDDRbase{Uniform} & \LDDRbase{72.6} & \LDDRbase{59.1} & \LDDRbase{50.0} & \LDDRbase{60.56} & \LDDRbase{69.3} & \LDDRbase{71.5} & \LDDRbase{54.6} & \LDDRbase{50.7} & \LDDRbase{57.22} & \LDDRbase{38.09} & \LDDRbase{59.03} \\
 &  & \LDDRbase{AKS} & \LDDRbase{72.8} & \LDDRbase{59.0} & \LDDRbase{51.1} & \LDDRbase{60.96} & \LDDRbase{69.3} & \LDDRbase{72.7} & \LDDRbase{60.0} & \LDDRbase{55.7} & \LDDRbase{61.11} & \LDDRbase{42.93} & \LDDRbase{63.19} \\
 &  & \LDDRbase{Q-frame} & \LDDRbase{74.4} & \LDDRbase{60.3} & \LDDRbase{52.0} & \LDDRbase{62.26} & \LDDRbase{69.3} & \LDDRbase{72.1} & \LDDRbase{56.6} & \LDDRbase{51.1} & \LDDRbase{58.04} & \LDDRbase{39.25} & \LDDRbase{62.81} \\
 &  & \LDDRbase{FOCUS} & \LDDRbase{72.0} & \LDDRbase{59.2} & \LDDRbase{52.2} & \LDDRbase{61.15} & \LDDRbase{68.3} & \LDDRbase{69.2} & \LDDRbase{60.9} & \LDDRbase{55.7} & \LDDRbase{60.81} & \LDDRbase{45.31} & \LDDRbase{63.25} \\
 &  & \LDDRbase{MDP3} & \LDDRbase{74.0} & \LDDRbase{61.7} & \LDDRbase{54.0} & \LDDRbase{63.22} & \LDDRbase{69.3} & \LDDRbase{72.3} & \LDDRbase{57.0} & \LDDRbase{50.9} & \LDDRbase{58.27} & \LDDRbase{39.25} & \LDDRbase{63.27} \\
\noalign{\vskip 2pt}\cdashline{3-14}[5pt/1.2pt]\noalign{\vskip 2pt}
\rowcolor{gray!10}
 &  & \LDDRbase{$\dagger$ \textbf{LD}} & \LDDRbase{74.2} & \LDDRbase{61.3} & \LDDRbase{53.0} & \LDDRbase{62.85\LDDRgain{2.3}} & \LDDRbase{67.7} & \LDDRbase{71.5} & \LDDRbase{63.4} & \LDDRbase{54.8} & \LDDRbase{\textbf{61.41}\LDDRgain{4.2}} & \LDDRbase{45.90\LDDRgain{7.8}} & \LDDRbase{\textbf{65.62}\LDDRgain{6.6}} \\
\rowcolor{gray!10}
\multirow{-14}{*}{\textbf{\shortstack{Qwen2.5-VL-7B*}}} &  & \LDDRbase{$\ddagger$ \textbf{LDDR}} & \LDDRbase{74.8} & \LDDRbase{63.7} & \LDDRbase{55.0} & \LDDRbase{\textbf{64.48}\LDDRgain{3.9}} & \LDDRbase{69.3} & \LDDRbase{70.3} & \LDDRbase{62.1} & \LDDRbase{54.4} & \LDDRbase{60.96\LDDRgain{3.7}} & \LDDRbase{\textbf{46.03}\LDDRgain{7.9}} & \LDDRbase{65.25\LDDRgain{6.2}} \\
\midrule
\multirow{7}{*}{\textbf{\shortstack{Qwen3-VL-8B*}}} & \multirow{7}{*}{8} & \LDDRbase{Uniform} & \LDDRbase{68.2} & \LDDRbase{53.6} & \LDDRbase{49.8} & \LDDRbase{57.19} & \LDDRbase{70.9} & \LDDRbase{66.9} & \LDDRbase{54.6} & \LDDRbase{45.2} & \LDDRbase{54.53} & \LDDRbase{28.08} & \LDDRbase{51.28} \\
 &  & \LDDRbase{AKS} & \LDDRbase{73.2} & \LDDRbase{59.1} & \LDDRbase{52.1} & \LDDRbase{61.48} & \LDDRbase{68.8} & \LDDRbase{72.1} & \LDDRbase{61.2} & \LDDRbase{51.4} & \LDDRbase{59.54} & \LDDRbase{40.15} & \LDDRbase{64.86} \\
 &  & \LDDRbase{Q-frame} & \LDDRbase{73.7} & \LDDRbase{59.4} & \LDDRbase{54.6} & \LDDRbase{62.56} & \LDDRbase{75.1} & \LDDRbase{72.7} & \LDDRbase{59.5} & \LDDRbase{49.8} & \LDDRbase{59.31} & \LDDRbase{35.05} & \LDDRbase{61.65} \\
 &  & \LDDRbase{FOCUS} & \LDDRbase{72.0} & \LDDRbase{58.8} & \LDDRbase{45.9} & \LDDRbase{58.89} & \LDDRbase{65.6} & \LDDRbase{75.0} & \LDDRbase{63.1} & \LDDRbase{50.2} & \LDDRbase{59.53} & \LDDRbase{31.57} & \LDDRbase{63.80} \\
 &  & \LDDRbase{MDP3} & \LDDRbase{73.0} & \LDDRbase{58.9} & \LDDRbase{54.1} & \LDDRbase{62.00} & \LDDRbase{73.5} & \LDDRbase{70.9} & \LDDRbase{60.9} & \LDDRbase{51.1} & \LDDRbase{59.83} & \LDDRbase{37.83} & \LDDRbase{68.76} \\
\noalign{\vskip 2pt}\cdashline{3-14}[5pt/1.2pt]\noalign{\vskip 2pt}
\rowcolor{gray!10}
 &  & \LDDRbase{$\dagger$ \textbf{LD}} & \LDDRbase{73.4} & \LDDRbase{62.3} & \LDDRbase{53.7} & \LDDRbase{63.15\LDDRgain{6.0}} & \LDDRbase{70.9} & \LDDRbase{73.3} & \LDDRbase{66.3} & \LDDRbase{55.9} & \LDDRbase{63.43\LDDRgain{8.9}} & \LDDRbase{43.06\LDDRgain{15.0}} & \LDDRbase{68.47\LDDRgain{17.2}} \\
\rowcolor{gray!10}
 &  & \LDDRbase{$\ddagger$ \textbf{LDDR}} & \LDDRbase{76.1} & \LDDRbase{62.4} & \LDDRbase{55.8} & \LDDRbase{\textbf{64.78}\LDDRgain{7.6}} & \LDDRbase{72.0} & \LDDRbase{77.3} & \LDDRbase{67.5} & \LDDRbase{56.2} & \LDDRbase{\textbf{64.62}\LDDRgain{10.1}} & \LDDRbase{\textbf{45.13}\LDDRgain{17.1}} & \LDDRbase{\textbf{70.58}\LDDRgain{19.3}} \\
\midrule
\multirow{7}{*}{\textbf{\shortstack{LLaVA-OneVision\\1.5-8B*}}} & \multirow{7}{*}{8} & \LDDRbase{Uniform} & \LDDRbase{67.1} & \LDDRbase{55.0} & \LDDRbase{48.7} & \LDDRbase{56.93} & \LDDRbase{65.1} & \LDDRbase{70.4} & \LDDRbase{54.9} & \LDDRbase{47.0} & \LDDRbase{54.97} & \LDDRbase{34.41} & \LDDRbase{57.10} \\
 &  & \LDDRbase{AKS} & \LDDRbase{68.8} & \LDDRbase{57.6} & \LDDRbase{49.7} & \LDDRbase{58.67} & \LDDRbase{65.1} & \LDDRbase{75.6} & \LDDRbase{61.9} & \LDDRbase{51.6} & \LDDRbase{59.76} & \LDDRbase{41.45} & \LDDRbase{63.60} \\
 &  & \LDDRbase{Q-frame} & \LDDRbase{67.9} & \LDDRbase{55.2} & \LDDRbase{50.0} & \LDDRbase{57.70} & \LDDRbase{63.5} & \LDDRbase{71.5} & \LDDRbase{57.0} & \LDDRbase{48.2} & \LDDRbase{56.10} & \LDDRbase{37.64} & \LDDRbase{58.79} \\
 &  & \LDDRbase{FOCUS} & \LDDRbase{68.0} & \LDDRbase{56.1} & \LDDRbase{47.0} & \LDDRbase{57.04} & \LDDRbase{61.9} & \LDDRbase{72.7} & \LDDRbase{62.9} & \LDDRbase{52.3} & \LDDRbase{59.54} & \LDDRbase{37.31} & \LDDRbase{64.17} \\
 &  & \LDDRbase{MDP3} & \LDDRbase{70.8} & \LDDRbase{56.4} & \LDDRbase{51.8} & \LDDRbase{59.66} & \LDDRbase{66.7} & \LDDRbase{74.4} & \LDDRbase{57.8} & \LDDRbase{51.1} & \LDDRbase{58.33} & \LDDRbase{40.02} & \LDDRbase{65.53} \\
\noalign{\vskip 2pt}\cdashline{3-14}[5pt/1.2pt]\noalign{\vskip 2pt}
\rowcolor{gray!10}
 &  & \LDDRbase{$\dagger$ \textbf{LD}} & \LDDRbase{70.8} & \LDDRbase{57.8} & \LDDRbase{52.2} & \LDDRbase{60.26\LDDRgain{3.3}} & \LDDRbase{65.1} & \LDDRbase{73.3} & \LDDRbase{60.0} & \LDDRbase{54.4} & \LDDRbase{60.06\LDDRgain{5.1}} & \LDDRbase{\textbf{44.87}\LDDRgain{10.5}} & \LDDRbase{66.49\LDDRgain{9.4}} \\
\rowcolor{gray!10}
 &  & \LDDRbase{$\ddagger$ \textbf{LDDR}} & \LDDRbase{71.3} & \LDDRbase{58.4} & \LDDRbase{53.0} & \LDDRbase{\textbf{60.93}\LDDRgain{4.0}} & \LDDRbase{65.6} & \LDDRbase{72.7} & \LDDRbase{61.9} & \LDDRbase{53.9} & \LDDRbase{\textbf{60.43}\LDDRgain{5.5}} & \LDDRbase{44.67\LDDRgain{10.3}} & \LDDRbase{\textbf{67.49}\LDDRgain{10.4}} \\
\midrule
\multirow{5}{*}{\textbf{\shortstack{InternVL3_5-4B}}} & \multirow{5}{*}{8} & \LDDRbase{Uniform} & \LDDRbase{69.7} & \LDDRbase{55.8} & \LDDRbase{47.6} & \LDDRbase{57.66} & \LDDRbase{63.5} & \LDDRbase{71.5} & \LDDRbase{55.1} & \LDDRbase{47.7} & \LDDRbase{55.27} & \LDDRbase{37.89} & \LDDRbase{61.38} \\
 &  & \LDDRbase{AKS} & \LDDRbase{70.1} & \LDDRbase{57.7} & \LDDRbase{50.2} & \LDDRbase{59.33} & \LDDRbase{67.2} & \LDDRbase{69.8} & \LDDRbase{58.7} & \LDDRbase{51.2} & \LDDRbase{58.18} & \LDDRbase{42.47} & \LDDRbase{65.87} \\
 &  & \LDDRbase{FOCUS} & \LDDRbase{68.0} & \LDDRbase{53.7} & \LDDRbase{45.1} & \LDDRbase{55.59} & \LDDRbase{58.7} & \LDDRbase{71.5} & \LDDRbase{59.2} & \LDDRbase{49.8} & \LDDRbase{56.76} & \LDDRbase{39.31} & \LDDRbase{66.25} \\
 &  & \LDDRbase{MDP3} & \LDDRbase{73.2} & \LDDRbase{57.4} & \LDDRbase{49.2} & \LDDRbase{59.96} & \LDDRbase{67.7} & \LDDRbase{72.1} & \LDDRbase{56.3} & \LDDRbase{48.6} & \LDDRbase{56.69} & \LDDRbase{42.93} & \LDDRbase{69.33} \\
\noalign{\vskip 2pt}\cdashline{3-14}[5pt/1.2pt]\noalign{\vskip 2pt}
\rowcolor{gray!10}
 &  & \LDDRbase{$\dagger$ \textbf{LD}} & \LDDRbase{72.4} & \LDDRbase{59.7} & \LDDRbase{50.4} & \LDDRbase{\textbf{60.85}\LDDRgain{3.2}} & \LDDRbase{64.6} & \LDDRbase{70.3} & \LDDRbase{61.9} & \LDDRbase{54.1} & \LDDRbase{\textbf{60.05}\LDDRgain{4.8}} & \LDDRbase{\textbf{46.28}\LDDRgain{8.4}} & \LDDRbase{68.98\LDDRgain{7.6}} \\
\midrule
\multirow{5}{*}{\textbf{\shortstack{InternVL3_5-8B}}} & \multirow{5}{*}{8} & \LDDRbase{Uniform} & \LDDRbase{68.6} & \LDDRbase{57.1} & \LDDRbase{50.4} & \LDDRbase{58.70} & \LDDRbase{66.7} & \LDDRbase{72.1} & \LDDRbase{54.6} & \LDDRbase{51.1} & \LDDRbase{57.06} & \LDDRbase{38.90} & \LDDRbase{61.34} \\
 &  & \LDDRbase{AKS} & \LDDRbase{72.8} & \LDDRbase{59.0} & \LDDRbase{52.0} & \LDDRbase{61.25} & \LDDRbase{66.7} & \LDDRbase{70.3} & \LDDRbase{61.7} & \LDDRbase{54.0} & \LDDRbase{60.20} & \LDDRbase{45.57} & \LDDRbase{66.23} \\
 &  & \LDDRbase{FOCUS} & \LDDRbase{69.6} & \LDDRbase{58.1} & \LDDRbase{46.3} & \LDDRbase{58.00} & \LDDRbase{56.1} & \LDDRbase{69.2} & \LDDRbase{60.9} & \LDDRbase{50.5} & \LDDRbase{56.91} & \LDDRbase{40.86} & \LDDRbase{65.99} \\
 &  & \LDDRbase{MDP3} & \LDDRbase{74.2} & \LDDRbase{59.2} & \LDDRbase{54.3} & \LDDRbase{62.59} & \LDDRbase{65.1} & \LDDRbase{70.9} & \LDDRbase{59.2} & \LDDRbase{51.1} & \LDDRbase{58.11} & \LDDRbase{43.51} & \LDDRbase{69.78} \\
\noalign{\vskip 2pt}\cdashline{3-14}[5pt/1.2pt]\noalign{\vskip 2pt}
\rowcolor{gray!10}
 &  & \LDDRbase{$\dagger$ \textbf{LD}} & \LDDRbase{72.3} & \LDDRbase{61.1} & \LDDRbase{52.9} & \LDDRbase{62.11\LDDRgain{3.4}} & \LDDRbase{67.2} & \LDDRbase{72.7} & \LDDRbase{61.7} & \LDDRbase{55.9} & \LDDRbase{\textbf{61.40}\LDDRgain{4.3}} & \LDDRbase{\textbf{48.22}\LDDRgain{9.3}} & \LDDRbase{\textbf{70.00}\LDDRgain{8.7}} \\
\bottomrule
\end{tabular}
\end{table*}

\textbf{Setting}: We applied a token-based formulation to control visual computation across all models. Following the resolution-to-token mapping of Qwen2.5-VL, each visual token corresponds to a $14 \times 14$ pixel patch before spatial merging. Thus, for a frame with spatial resolution $H \times W$, its visual-token count is computed as $w = \frac{H \times W}{14^2}.$ In our experiments, we treat the per-frame token count $w_t$ as the primary control variable and constrain it within $w_{\min} \leq w_t \leq w_{\max}$, where $w_{\min}=256$ and $w_{\max}=1024$. Given a target token count $w_t$, we derive the corresponding frame resolution by preserving the original aspect ratio and projecting the resized frame to the nearest valid patch grid. This token--resolution mapping is derived from Qwen2.5-VL and is applied consistently across all evaluated models. To ensure fair comparison, we define a visual token budget using a frame-equivalent formulation: $C_{\mathrm{total}} = F \times 1024$, where $F$ denotes the frame-equivalent budget and 1024 tokens correspond to one maximum-resolution frame. Under fixed-resolution settings, each selected frame uses $w_t=1024$ tokens, so exactly $|\mathcal S|=K=F$ frames are selected. Under dynamic-resolution settings, LDDR first constructs a larger candidate set $\mathcal S$ with $|\mathcal S|=K=2F$ using LD, and then applies GD-guided sorted-prefix retention and token allocation to obtain the final set $\mathcal S^*$ of size $k^*$. Here, $k^*$ is typically larger than $F$, since lower-resolution frames allow more frames to be retained within the same budget. Frames in $\mathcal S^*$ are assigned variable token counts while satisfying $\sum_{t\in\mathcal S^*} w_t \leq C_{\mathrm{total}}$ and $w_t \in [256,1024]$. (guaranteeing $\leq C_{\mathrm{total}}$ visual tokens, excluding \texttt{|vision\_start|} and \texttt{|vision\_end|} formatting)

\section{Results}
We follow the baselines described in \S \ref{sec:experiments} and evaluate frame budgets ranging from 8 to 32 frames.
Table~\ref{tab:main_results_full} shows that LD and LDDR consistently outperform most baselines across five models and four benchmarks. Averaging the 8-frame setting and 32-frame setting in Qwen2.5-VL, LDDR beats the best baseline by 2.5 and 1.6 respectively. These gains are more pronounced in longer videos. In the 3600s subset of LongVideoBench, LD's performance reaches 54.8 (a 2-point gain over the best baseline), indicating that \textbf{global-wise Linear DPP selection effectively captures long-range dependencies and removes redundancy}. Notably, dynamic resolution improves upon fixed-resolution LD
in 12 out of 16 evaluated settings, \textbf{with the largest
additional gains observed under constrained frame budgets}.

\begin{table}[t]
\centering
\caption{Qwen2.5-VL-32B results on LongVideoBench under 64-frame-equivalent budgets.}
\label{tab:large_model_budget}
\small
\setlength{\tabcolsep}{3pt}
\renewcommand{\arraystretch}{0.7}
\begin{tabular}{clccccc}
\toprule
\#F & Method & 15s & 60s & 600s & 3600s & Overall \\
\midrule
\multirow{3}{*}{64}
& Uniform & 75.13 & 71.51 & 63.11 & 56.91 & 63.27 \\
& $\dagger$ LD      & 70.90 & 72.67 & 67.96 & 59.57 & 65.45 \\
\rowcolor{gray!15}
& $\ddagger$LDDR    & 71.95 & 70.93 & 67.48 & 60.64
& \textbf{65.67} \\
\bottomrule
\end{tabular}
\end{table}

\subsection{Scaling to Larger Models and Frame Budgets}
\label{sec:budget_scaling}

To evaluate LDDR under larger budgets, we use Qwen2.5-VL-32B, which supports longer contexts. We evaluate LDDR on LongVideoBench with a 64-frame-equivalent budget. As shown in Table~\ref{tab:large_model_budget},
both LD and LDDR maintain their advantage over uniform sampling as the visual budget increases.

\subsection{Analysis 1: Efficiency Improvement with Linear DPP}

\begin{figure}[htbp]
        \centering
        \includegraphics[width=0.48\textwidth]{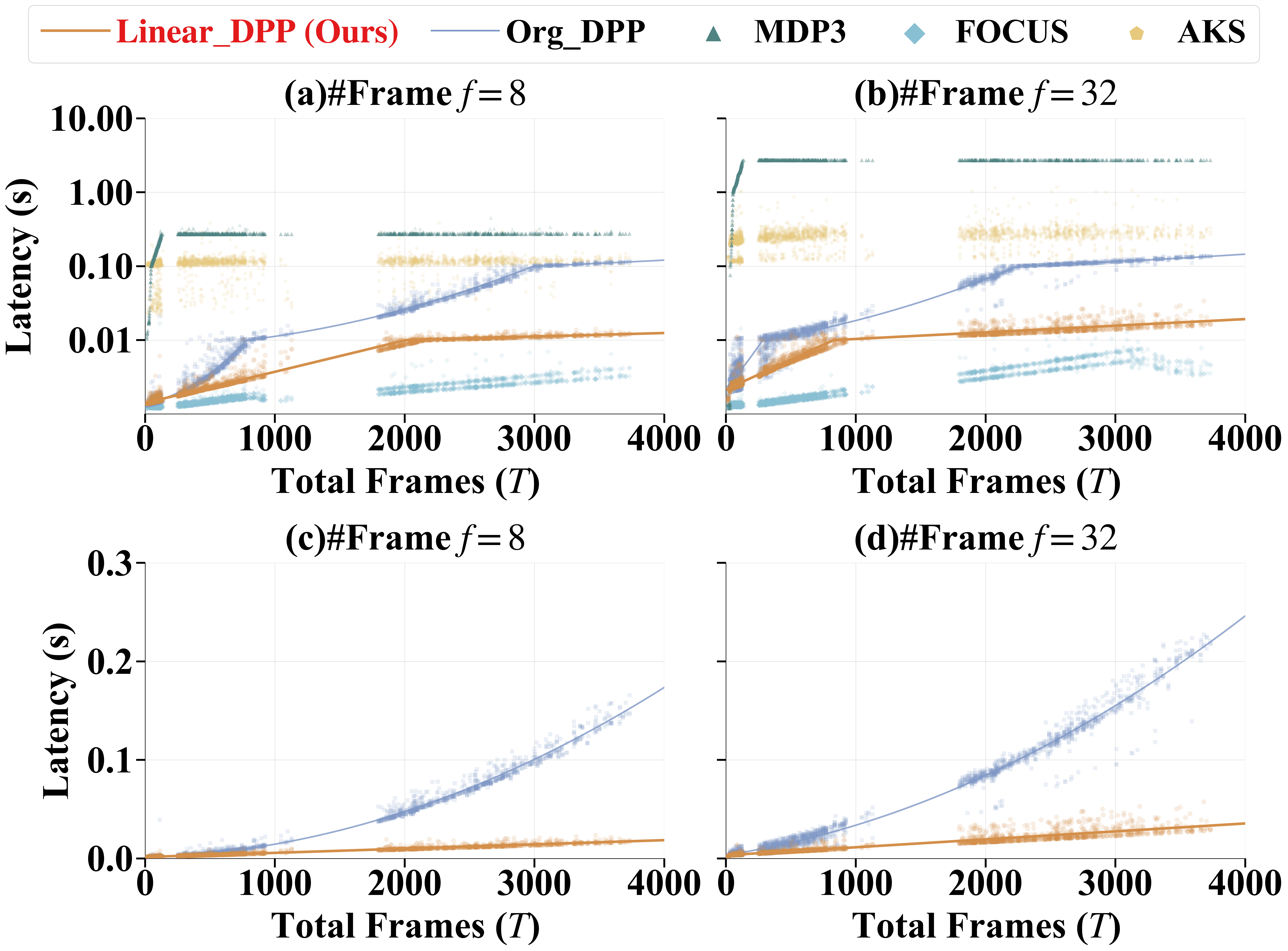}
 
    \caption{Sampling runtime under different total numbers of input frames on Video-MME. Up: comparison across different frame sampling methods. Down: comparison between Linear DPP implementation and the original DPP. The runtime of Linear DPP scales approximately linearly as the total number of frames increases.}
    \label{fig:efficiency_all}
\end{figure}

\begin{figure}[htbp]
    \centering
    \captionsetup{font=small}
    \includegraphics[width=\linewidth]{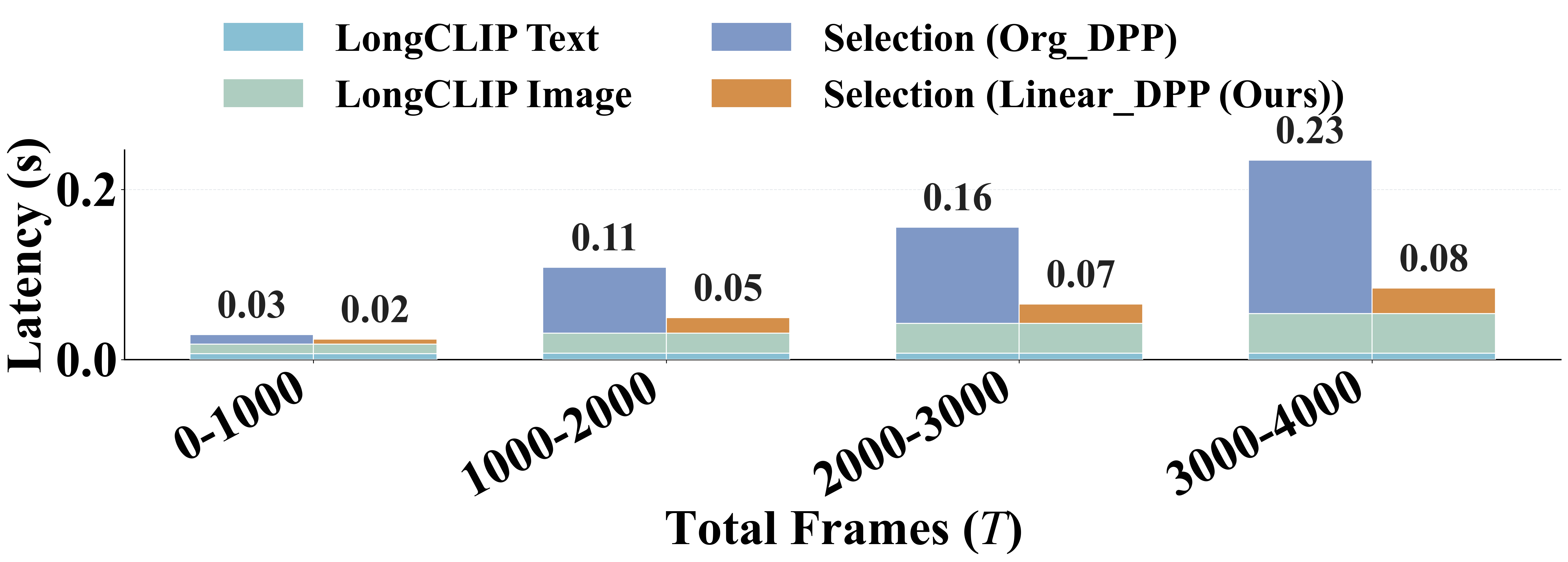}
    \caption{Latency breakdown by phase, including LongCLIP processing and frame sampling.}
    \label{fig:efficiency_each_phase}
\end{figure}

End-to-end latency is critical for MLLM video reasoning, while modern inference engines have reduced the time to first token (TTFT) of 8B-scale language models to hundreds of milliseconds; frame sampling can introduce substantial latency before model inference begins. Therefore, optimizing frame sampling time is useful. LD remains efficient by performing selection directly in the task-conditioned feature space. Shown in Figure~\ref{fig:efficiency_all}, kernel-space scales quadratically with the length, while LD follows the proposed formulation $\mathcal{O}(KTd)$ and exhibits quasi-linear scaling. Figure~\ref{fig:efficiency_each_phase} further proves that LD mainly reduces selection overhead, the dominant bottleneck in video pre-processing. \textbf{LD makes global-wise frame selection efficient for long videos and achieves the lowest inference latency among AKS and MDP3.}

\subsection{Analysis 2: Global Selection vs. Chunked Selection}

The linear feature-space formulation allows LD to optimize frame selection over the entire video, whereas several prior methods like \citep{sun2025mdp3} only perform chunk-level selection.

Table~\ref{tab:chunk_ablation} compares full-video selection with chunked variants under the same frame budget, with the budget evenly distributed across chunks. The global setting achieves the best overall score of 61.4, while splitting the video into 2 chunks reduces the score to 59.6. The degradation is most pronounced on long-duration 600s and 3600s subsets. This is because chunking imposes fixed local budgets, which can waste frames on low-information segments while preventing query-relevant frames in distant segments from competing globally. \textbf{These results support the importance of global selection for long-video understanding.}

\begin{table}[htbp]
\centering
\caption{Analysis on chunked sampling. ``Global'' denotes our LD sampling over the entire video, while $k_{\scriptsize(n\times k)}$ denotes dividing the video into $k$ chunks and sampling $n$ frames by LD from each chunk.}
\label{tab:chunk_ablation}
\small
\setlength{\tabcolsep}{3pt}
\renewcommand{\arraystretch}{0.6}
\begin{tabular}{ccccccc}
\toprule
\multicolumn{7}{c}{LongVideoBench} \\
\midrule
\#F & Chunk & 15s & 60s & 600s & 3600s & Overall \\
\midrule
\multirow{5}{*}{32}
& Global & 67.7 & 71.5 & 63.4 & 54.8 & \textbf{61.4} \\
& $2_{\scriptsize(16\times2)}$
& 67.7 & 72.7 & 60.7 & 52.1 & 59.6 \\
& $4_{\scriptsize(8\times4)}\phantom{1}$
& 67.7 & 73.3 & 59.0 & 52.8 & 59.5 \\
& $8_{\scriptsize(4\times8)}\phantom{1}$
& 67.7 & 73.3 & 59.0 & 52.8 & 59.5 \\
& $16_{\scriptsize(2\times16)}$
& 67.7 & 72.7 & 58.3 & 53.0 & 59.2 \\
\bottomrule
\end{tabular}
\end{table}

\newcommand{\sn}[1]{{\scriptsize #1}}
\newcolumntype{s}{>{\scriptsize}c}

\section{Ablation Study}

\subsection{Ablation 1: GD Importance Scoring and Dynamic Resolution Strategy}

Table~\ref{tab:dr_ablation} ablates the GD importance scoring and dynamic-resolution allocation used in LDDR.
All variants are evaluated under the same overall visual token budget. We compare LDDR with fixed-resolution baselines, where all selected frames use the same token budget ($w_t=1024$ (high), $512$ (med), or $256$ (low)), and a hand-crafted multi-resolution baseline ($4\times1024+4\times512+8\times256$) inspired by~\citet{zhang2025q}. For hand-crafted multi-resolution allocation, token assignment is determined by the scoring sequence. To isolate the effect of the dynamic resolution scoring in Eq. \ref{eq:GD}, we replace GD importance with simple relevance scoring. The LDDR achieves the best overall performance. Fixed-resolution sampling performs worse. The hand-crafted multi-resolution variant improves over fixed allocation, suggesting that resolution diversity is useful. Replacing GD with relevance-only scoring also degrades performance, indicating that GD scoring can better attribute the frame's marginal contribution to the candidate frame set. \textbf{Results show that our GD importance scoring and dynamic resolution strategy are both effective.}
\begin{table}[htbp]
\centering
\caption{Ablation on scoring and resolution strategies; \textbf{Rel} denotes scoring frame importance by cosine similarity between frame and query. All scoring and dynamic resolution are performed after the candidate set $S$ obtained by LD.
First row: Our full LDDR. Upper part: different resolution strategies; Lower part: different importance scoring mechanisms.}
\label{tab:dr_ablation}
\scriptsize
\setlength{\tabcolsep}{2.5pt}
\renewcommand{\arraystretch}{0.8}

\begin{tabular}{cc|cc|ccccc}
\toprule
\multicolumn{2}{c|}{Frame Scoring}
& \multicolumn{2}{c|}{Frame Resolution}
& \multicolumn{5}{c}{LongVideoBench} \\
\cmidrule(lr){1-2}
\cmidrule(lr){3-4}
\cmidrule(lr){5-9}
\textbf{GD (Ours)} & \textbf{Rel}
& \textbf{DR (Ours)} & \textbf{Hard}
& 15s & 60s & 600s & 3600s & Overall \\
\midrule
\rowcolor{gray!15}
\cmark & \xmark & \cmark & \xmark
& 69.3 & 73.3 & 60.7 & 55.9 & \textbf{61.48} \\
- & - & \xmark & \cmark\text{-high}
& 66.7 & 69.8 & 59.2 & 54.8 & 59.76 \\
- & - & \xmark & \cmark\text{-med}
& 67.2 & 68.0 & 61.6 & 56.4 & 61.03 \\
- & - & \xmark & \cmark\text{-low}
& 68.3 & 70.3 & 59.7 & 52.7 & 59.31 \\
\cmark & \xmark & \xmark & \cmark\text{-multi}
& 70.4 & 72.1 & 58.0 & 57.1 & 61.18 \\\midrule
\xmark & \cmark & \xmark & \cmark\text{-multi}
& 69.3 & 70.9 & 60.7 & 54.6 & 60.65 \\
\xmark & \cmark & \cmark & \xmark
& 66.1 & 72.7 & 60.2 & 55.5 & 60.66 \\
\bottomrule
\end{tabular}
\end{table}

\subsection{Ablation 2: Diversity vs. Relevance, Encoder Sensitivity and Density Prior}
We ablate the DPP components in Table~\ref{tab:coverage_relevance}, where 
$L_{ij}=r_i\phi_{ij}r_j$ contains query-conditioned relevance $r_i$ and similarity $\phi_{ij}$ preserved diversity. 
The full setting performs best at 59.96; removing relevance drops performance to 51.37, while removing diversity reduces it to 54.89, confirming that relevance identifies informative frames and diversity suppresses redundancy. 
We further compare CLIP variants in Table~\ref{tab:clip_variants}. 
LongCLIP performs best. Since LDDR performs global selection throughout the video, better visual features enable more effective selection. We also include a detailed sensitivity analysis of the density prior exponent $\tau$ in Appendix~\ref{density-prior}.
\section{Qualitative Analysis}
\subsection{Same Video, Different Questions: Different Frame Selections and Token Allocations}
\begin{figure}[htbp]
    \centering
    \includegraphics[width=0.5\textwidth]{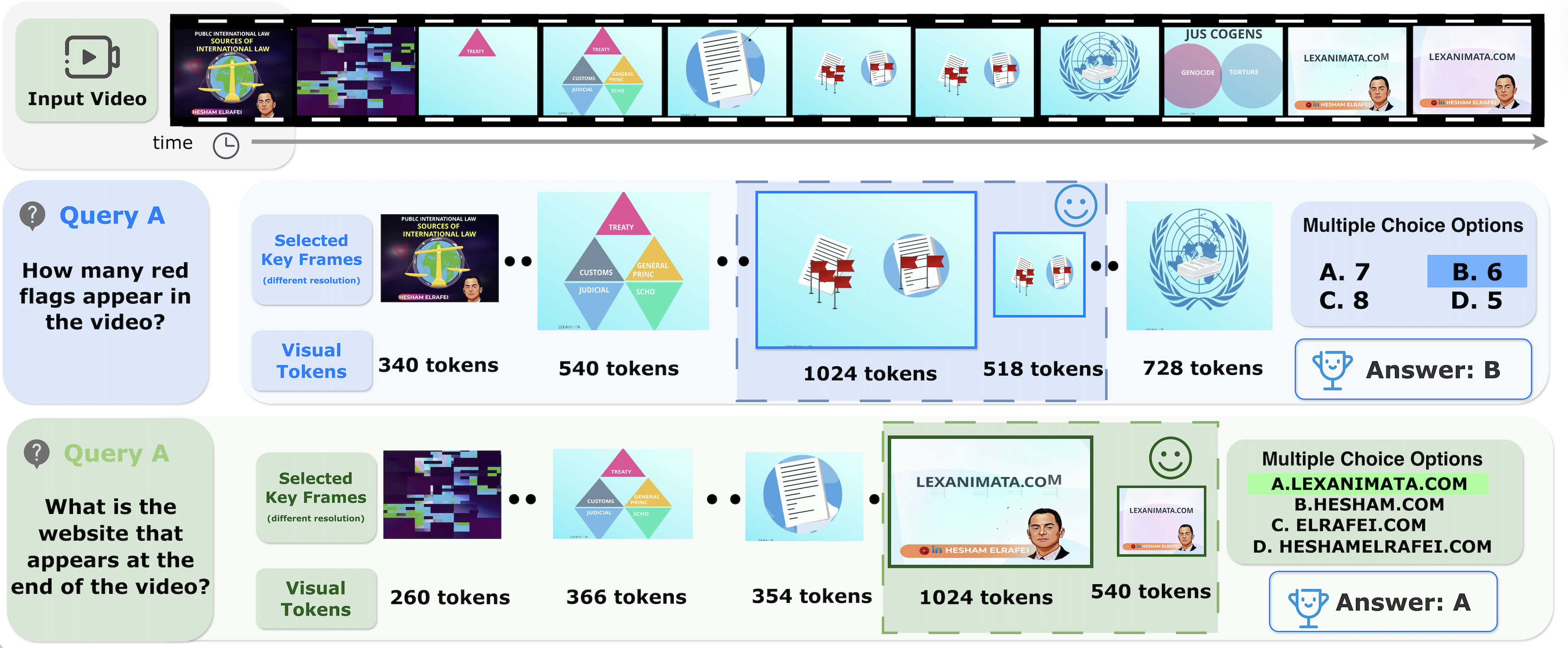}
    \caption{Qualitative examples of LDDR on video question answering. Same video with different questions.}
    \label{fig:my_imag1}
\end{figure}
Figure~\ref{fig:my_imag1} provides an example where different user questions are asked on a same video, which shows how LDDR can wisely select and allocate frame budgets. In the first counting problem, LDDR captures distinct appearances of the red flags. Dynamic resolution further improves efficiency by assigning more visual tokens to important frames with flag details, while avoiding repeatedly allocating high token budgets to similar frames. In the second example, although it is the same video, different questions lead to different selections. When the final website text ``LEXANIMATA.COM'' appears in consecutive frames, LDDR assigns a high token budget to the most informative frame. These examples show that global diversity, query relevance, and adaptive token allocation retain a compact yet informative video representation.

\subsection{Failure Case Analysis}
\label{sec:failure_analysis}
We also manually inspect 50 cases where LD answers correctly, but LDDR does not. In 29 cases, the chosen frames contain required evidence, but due to the limited ability of MLLM, it fails to answer it correctly. Another 15 cases exhibit potentially insufficient resolution, while 6 cases lack critical evidence or the temporal context. Representative cases are further discussed in Appendix~\ref{appendix:failure_cases}.

\section{Conclusion}
LDDR is a training-free framework for efficient long-video understanding. It performs global query-aware DPP selection in linear space and with Group DPP importance to wisely allocate visual tokens.
Across multiple benchmarks and MLLM backbones, LDDR outperforms uniform sampling and most existing selection methods. Especially showing high potential on long videos and constrained budgets. Ablations confirm the benefits of global selection, diversity-aware scoring, and dynamic resolution. Qualitative analysis further demonstrates the method's effectiveness in real use cases, making LDDR a practical frame selection and token allocation strategy for video MLLMs. 

\section*{Limitations}
Although LDDR improves the efficiency of long-video understanding by selecting informative frames and assigning dynamic resolution, several limitations remain. First, LDDR relies on text-visual features extracted from external encoders, so its performance can be affected by the quality of that external module. Second, LDDR is designed as a training-free preprocessing method for existing MLLM backbones, and therefore, its final performance still depends on the reasoning capacity, ability to handle dynamic resolution, and visual-token handling capability of the downstream model.

\section*{Code}
Code Link: https://github.com/JingfengChen-Jay/LDDR

\bibliography{custom}

\appendix

\section{Effect of density-prior}
\label{density-prior}

\begin{table*}[t]
\centering
\caption{
Ablation study on the density-prior exponent $\tau$.
$\tau=0$ removes the density prior, while $\tau=1$ is used as the default setting in all main experiments.
}
\label{tab:alpha_ablation}
\small
\setlength{\tabcolsep}{4pt}
\renewcommand{\arraystretch}{1.1}
\begin{tabular}{c cccc ccccc}
\toprule
\multirow{2}{*}{$\tau$}
& \multicolumn{4}{c}{\textbf{Video-MME}}
& \multicolumn{5}{c}{\textbf{LongVideoBench}} \\
\cmidrule(lr){2-5} \cmidrule(lr){6-10}
& Short & Med & Long & Overall
& 15s & 60s & 600s & 3600s & Overall \\
\midrule
0.0   & 73.11 & 58.11 & 52.67 & 61.29 & 66.14 & 70.93 & 61.65 & \textbf{57.09} & 61.55 \\
0.3 & 72.56 & 58.89 & 53.11 & 61.52 & \textbf{71.43} & 71.51 & 61.17 & 56.56 & 62.00 \\
0.5 & \textbf{73.44} & 58.67 & \textbf{53.33} & \textbf{61.81} & \textbf{71.43} & \textbf{73.26} & 59.95 & 56.91 & \textbf{62.01} \\
0.7 & 72.56 & 58.89 & 52.78 & 61.41 & 69.31 & 72.09 & \textbf{61.41} & 56.21 & 61.71 \\
1.0 default & 72.89 & \textbf{59.78} & 52.78 & 61.82 & 69.31 & \textbf{73.26} & 60.68 & 55.85 & 61.48 \\
\bottomrule
\end{tabular}
\end{table*}
Table~\ref{tab:alpha_ablation} studies the effect of the density-prior exponent $\tau$ in
$\tilde{\mathcal I}_t=\mathcal I_t\rho_t^\tau$.
The comparison between $\tau=0$ and $\tau>0$ confirms the effectiveness of the density prior: incorporating relevance-weighted feature density consistently improves or maintains the overall performance.
Empirically, $\tau=0.5$ gives the strongest average result among the tested settings, especially on LongVideoBench.
Nevertheless, we intentionally avoid selecting $\tau$ based on these test-set results.
Such tuning would make the method dependent on the target benchmark and compromise the fairness of comparison.
Accordingly, we use the simple default $\tau=1$ for all main experiments, while reporting the full ablation to show that the method is robust to different density-prior strengths.

\section{Failure Cases:}
\label{appendix:failure_cases}
\begin{tcolorbox}[
    title={Case 1: Insufficient Temporal Context},
    colback=gray!15,
    colframe=black!70,
    boxrule=0.5pt
]
\textbf{Benchmark:} LongVideoBench\\
\textbf{Sample:} \texttt{0\_YDrJoUe8s\_1}\\
\textbf{Failure type:} Critical evidence not retained.

\medskip
\textbf{Observation.}
LDDR retains the queried visual change but fails to
preserve sufficient surrounding temporal context to
associate it with the referenced utterance.
Consequently, LDDR predicts an incorrect answer,
whereas LD answers correctly.

\medskip
\textbf{Interpretation.}
Although the relevant visual event is retained,
the contextual evidence required to interpret it is
incomplete. This case illustrates that query-relevant
frame selection alone may not preserve the temporal
relationships needed for correct reasoning.
\end{tcolorbox}

\begin{tcolorbox}[
    title={Case 2: Resolution-Related Ambiguity},
    colback=gray!15,
    colframe=black!70,
    boxrule=0.5pt
]
\textbf{Benchmark:} Video-MME\\
\textbf{Sample:} \texttt{001-1}\\
\textbf{Failure type:} Potentially insufficient spatial
resolution.

\medskip
\textbf{Observation.}
LDDR retains the decisive frame showing that berries
are the most numerous decoration and assigns it the
highest resolution among its retained frames.
Nevertheless, LDDR answers incorrectly, whereas LD
answers correctly. The decisive frame remains
lower-resolution than its counterpart in LD.

\medskip
\textbf{Interpretation.}
The error may arise from reduced fine-grained visual
detail or interference from additional temporal
context. These factors cannot be isolated from the
observed predictions alone, so we conservatively
categorize this case as potentially resolution-related.
\end{tcolorbox}

These cases highlight the trade-off between broader
temporal coverage, contextual completeness, and
per-frame spatial resolution under a fixed visual-token
budget. Retaining a relevant frame does not necessarily
ensure that sufficient context and visual detail remain
available for correct downstream reasoning.

\section{Diversity vs. Relevance and Encoder Sensitivity}
Table~\ref{tab:clip_and_component} contains DPP components
ablation and Encoder Sensitivity Analysis.
\begin{table*}[htbp]
\centering
\small
\setlength{\tabcolsep}{4pt}
\caption{
Component and encoder ablations on Video-MME. 
Left: removing either relevance or diversity substantially degrades performance. 
Right: LongCLIP achieves the best performance, showing that stronger vision-language alignment improves frame selection.
}
\renewcommand{\arraystretch}{1.05}
\begin{minipage}[t]{0.35\linewidth}
\centering
\begin{subtable}[t]{\linewidth}
\caption{Ablation on two components of DPP: Diversity and Relevance.}
\label{tab:coverage_relevance}
\begin{tabular}{l c c c c}
\toprule
Method & Short & Med & Long & Overall \\
\midrule
LD            & \textbf{73.30} & \textbf{57.00} & \textbf{49.56} & \textbf{59.96} \\
w/o Relevance  & 61.11          & 47.67          & 45.33          & 51.37          \\
w/o Diversity   & 65.00          & 52.89          & 46.78          & 54.89          \\
\bottomrule
\end{tabular}
\end{subtable}
\end{minipage}
\hfill
\begin{minipage}[t]{0.45\linewidth}
\centering
\begin{subtable}[t]{\linewidth}
\caption{Sensitivity Analysis of different text-visual encoders.}
\label{tab:clip_variants}

\begin{tabular}{c c c c c}
\toprule
  Encoder & Short & Med & Long & Overall \\
\midrule
 CLIP     & 64.33 & 49.22 & 48.00 & 53.85 \\
 SigLIP   & 67.88 & 52.11 & 47.33 & 55.78 \\
 LongCLIP & \textbf{73.30} & \textbf{57.00} & \textbf{49.56} & \textbf{59.96} \\
\bottomrule
\end{tabular}
\end{subtable}
\end{minipage}
\label{tab:clip_and_component}
\end{table*}

\section{Equivalence Between Standard Greedy MAP and Kernel-Free Greedy MAP}
\label{appendix:equivalence}

In this section, we prove that the proposed kernel-free DPP inference is equivalent to standard greedy MAP inference applied to the modulated kernel
\[
\tilde{\mathbf L}=\mathrm{diag}(\tilde{\mathbf r})\,\mathbf S\,\mathrm{diag}(\tilde{\mathbf r}),
\]
rather than an approximation. Therefore, both methods optimize the same objective and produce the same greedy selection sequence under exact arithmetic and identical tie-breaking.


\paragraph{Step 1: Factorization of the modulated kernel.}
Let
\[
\hat{\mathbf F}=
\begin{bmatrix}
\hat{\mathbf f}_1^\top\\
\vdots\\
\hat{\mathbf f}_T^\top
\end{bmatrix}
\in\mathbb R^{T\times d},
\qquad
\hat{\mathbf f}_t=\frac{\mathbf h_t}{\|\mathbf h_t\|}.
\]
Since
\[
\mathbf S=\hat{\mathbf F}\hat{\mathbf F}^\top,
\]
the modulated kernel can be written as
\begin{equation}
\tilde{\mathbf L}
=
\mathrm{diag}(\tilde{\mathbf r})\,\hat{\mathbf F}\hat{\mathbf F}^\top\,\mathrm{diag}(\tilde{\mathbf r}).
\end{equation}
Define
\begin{equation}
\mathbf \Phi := \mathrm{diag}(\tilde{\mathbf r})\,\hat{\mathbf F}\in\mathbb R^{T\times d},
\end{equation}
whose $t$-th row is denoted by $\boldsymbol\phi_t^\top\in\mathbb R^d$. Then
\begin{equation}
\tilde{\mathbf L}=\mathbf \Phi \mathbf \Phi^\top,
\qquad
\tilde L_{st}=\boldsymbol\phi_s^\top \boldsymbol\phi_t.
\label{eq:L_factorization}
\end{equation}
Hence the modulated DPP kernel is exactly a Gram matrix in the task-conditioned feature space.
\paragraph{Step 2: Greedy MAP objective.}
For an $L$-ensemble DPP, the MAP objective is
\begin{equation}
\mathcal S^*=\arg\max_{\mathcal S\subseteq [T]} \det(\tilde{\mathbf L}_{\mathcal S}),
\label{eq:dpp_map_obj}
\end{equation}
where $\tilde{\mathbf L}_{\mathcal S}$ is the principal submatrix indexed by $\mathcal S$ where $|\mathcal S|=K$.
Using~\eqref{eq:L_factorization},
\[
\tilde{\mathbf L}_{\mathcal S}=\mathbf \Phi_{\mathcal S}\mathbf \Phi_{\mathcal S}^\top.
\]
Therefore,
\begin{equation}
\det(\tilde{\mathbf L}_{\mathcal S})
=
\det(\mathbf \Phi_{\mathcal S}\mathbf \Phi_{\mathcal S}^\top),
\label{eq:det_equivalence}
\end{equation}
which shows that standard DPP MAP on $\tilde{\mathbf L}$ and feature-space DPP MAP on $\mathbf \Phi$ optimize the \emph{same determinant objective}. The only difference is whether the computation is carried out in the $T$-dimensional sample space or in the $d$-dimensional feature space.

\paragraph{Step 3: Standard greedy MAP in sample space.}
Let $d_t^{(i)}$ denote the marginal gain residual after $i$ greedy steps. Standard Cholesky-style greedy MAP inference on $\tilde{\mathbf L}$ initializes
\begin{equation}
d_t^{(0)}=\tilde L_{tt}.
\end{equation}
At step $i$, suppose the selected item is
\[
j_i=\arg\max_t d_t^{(i-1)}.
\]
Then the new basis vector $\mathbf e_i\in\mathbb R^T$ is computed as
\begin{equation}
\mathbf e_i
=
\frac{
\tilde{\mathbf L}_{:,j_i}
-
\sum_{k=1}^{i-1} e_{k,j_i}\,\mathbf e_k
}{
\sqrt{d_{j_i}^{(i-1)}}},
\label{eq:standard_e_update}
\end{equation}
and the residual gains are updated by
\begin{equation}
d_t^{(i)}=d_t^{(i-1)}-e_{i,t}^2.
\label{eq:standard_d_update}
\end{equation}
Since $\tilde{\mathbf L}$ is symmetric, the row/column form is equivalent.

\paragraph{Step 4: Kernel-free greedy MAP in feature space.}
The proposed method instead maintains basis vectors $\mathbf c_i\in\mathbb R^d$ in the feature space:
\begin{equation}
\mathbf c_i
=
\frac{
\boldsymbol\phi_{j_i}
-
\sum_{k=1}^{i-1}
(\mathbf c_k^\top \boldsymbol\phi_{j_i})\,\mathbf c_k
}{
\sqrt{d_{j_i}^{(i-1)}}},
\label{eq:feature_c_update}
\end{equation}
with gain update
\begin{equation}
d_t^{(i)}
=
d_t^{(i-1)}-(\boldsymbol\phi_t^\top \mathbf c_i)^2.
\label{eq:feature_d_update}
\end{equation}
In matrix form, this is
\[
\mathbf d^{(i)}=\mathbf d^{(i-1)}-(\mathbf \Phi \mathbf c_i)^2,
\]
where the square is elementwise.
\paragraph{Theorem 1.}
\emph{For the kernel factorization $\tilde{\mathbf L}=\mathbf \Phi \mathbf \Phi^\top$, the sample-space greedy updates~\eqref{eq:standard_e_update}--\eqref{eq:standard_d_update} and the feature-space greedy updates~\eqref{eq:feature_c_update}--\eqref{eq:feature_d_update} are exactly equivalent. In particular, for every iteration $i$,}
\begin{equation}
\mathbf e_i=\mathbf \Phi \mathbf c_i,
\label{eq:e_equals_Phic}
\end{equation}
\emph{and both methods maintain identical residual gains $d_t^{(i)}$. Consequently, they choose the same greedy item $j_i$ at every step.}
\paragraph{Proof.}
We prove the statement by induction.

\medskip
\noindent\textbf{Base case.}
At initialization,
\[
d_t^{(0)}=\tilde L_{tt}.
\]
Using $\tilde{\mathbf L}=\mathbf \Phi\mathbf \Phi^\top$,
\begin{equation}
d_t^{(0)}=\tilde L_{tt}
=\boldsymbol\phi_t^\top \boldsymbol\phi_t
=\|\boldsymbol\phi_t\|_2^2.
\label{eq:base_gain}
\end{equation}
Thus both methods start from the same gain vector and therefore select the same first item
\[
j_1=\arg\max_t d_t^{(0)},
\]
assuming identical tie-breaking.

For the first selected item, the standard sample-space update gives
\[
\mathbf e_1
=
\frac{\tilde{\mathbf L}_{:,j_1}}
{\sqrt{d_{j_1}^{(0)}}}.
\]
Since $\tilde{\mathbf L}=\mathbf \Phi\mathbf \Phi^\top$, we have
\[
\tilde{\mathbf L}_{:,j_1}
=
\mathbf \Phi \boldsymbol\phi_{j_1}.
\]
On the other hand, the feature-space update at the first step is
\[
\mathbf c_1
=
\frac{\boldsymbol\phi_{j_1}}
{\sqrt{d_{j_1}^{(0)}}}.
\]
Therefore,
\[
\mathbf e_1
=
\frac{\mathbf \Phi \boldsymbol\phi_{j_1}}
{\sqrt{d_{j_1}^{(0)}}}
=
\mathbf \Phi
\frac{\boldsymbol\phi_{j_1}}
{\sqrt{d_{j_1}^{(0)}}}
=
\mathbf \Phi \mathbf c_1.
\]
Hence the base case establishes both identical initial gains and
$\mathbf e_1=\mathbf \Phi \mathbf c_1$.

\medskip
\noindent\textbf{Induction hypothesis.}
Assume that after $i-1$ steps, both methods have identical gains $d_t^{(i-1)}$, hence they choose the same next item
\[
j_i=\arg\max_t d_t^{(i-1)}.
\]
Assume also that for all $k<i$,
\[
\mathbf e_k=\mathbf \Phi \mathbf c_k.
\]

\medskip
\noindent\textbf{Induction step: equivalence of the new basis vector.}
From the standard sample-space update~\eqref{eq:standard_e_update},
\[
\mathbf e_i
=
\frac{
\tilde{\mathbf L}_{:,j_i}
-
\sum_{k=1}^{i-1} e_{k,j_i}\,\mathbf e_k
}{
\sqrt{d_{j_i}^{(i-1)}}}.
\]
Now use $\tilde{\mathbf L}=\mathbf \Phi\mathbf \Phi^\top$. The $j_i$-th column is
\[
\tilde{\mathbf L}_{:,j_i}=\mathbf \Phi \boldsymbol\phi_{j_i}.
\]
Also, by the induction hypothesis,
\[
e_{k,j_i}
=
(\mathbf e_k)_{j_i}
=
(\mathbf \Phi \mathbf c_k)_{j_i}
=
\boldsymbol\phi_{j_i}^\top \mathbf c_k.
\]
Substituting these into the sample-space update gives
\begin{align}
\mathbf e_i
&=
\frac{
\mathbf \Phi \boldsymbol\phi_{j_i}
-
\sum_{k=1}^{i-1}
(\boldsymbol\phi_{j_i}^\top \mathbf c_k)\,
\mathbf \Phi \mathbf c_k
}{
\sqrt{d_{j_i}^{(i-1)}}}
\\
&=
\mathbf \Phi
\frac{
\boldsymbol\phi_{j_i}
-
\sum_{k=1}^{i-1}
(\mathbf c_k^\top \boldsymbol\phi_{j_i})\,\mathbf c_k
}{
\sqrt{d_{j_i}^{(i-1)}}}.
\end{align}
The term in parentheses is exactly the feature-space update~\eqref{eq:feature_c_update}, so
\[
\mathbf e_i=\mathbf \Phi \mathbf c_i.
\]
This proves~\eqref{eq:e_equals_Phic}.

\medskip
\noindent\textbf{Induction step: equivalence of gain updates.}
Using~\eqref{eq:e_equals_Phic},
\[
e_{i,t}=(\mathbf \Phi \mathbf c_i)_t=\boldsymbol\phi_t^\top \mathbf c_i.
\]
Therefore the standard residual update~\eqref{eq:standard_d_update} becomes
\[
d_t^{(i)}
=
d_t^{(i-1)}-e_{i,t}^2
=
d_t^{(i-1)}-(\boldsymbol\phi_t^\top \mathbf c_i)^2,
\]
which is exactly the kernel-free update~\eqref{eq:feature_d_update}. Hence both methods maintain the same gain vector after step $i$.

By induction, the equivalence holds for all iterations. Since the gain vectors are identical at every step, both algorithms choose the same next item $j_i$ and therefore produce the same greedy sequence and the same final subset.

\paragraph{Interpretation.}
The standard greedy algorithm orthogonalizes columns of $\tilde{\mathbf L}$ in the $T$-dimensional sample space, whereas the kernel-free algorithm orthogonalizes row features $\boldsymbol\phi_t$ in the $d$-dimensional latent space. Theorem~1 shows that these are two algebraically equivalent views of the same Cholesky/Gram--Schmidt process.

\paragraph{Corollary 1 (Objective consistency).}
\emph{The proposed kernel-free inference does not change the DPP objective. It maximizes the same greedy surrogate of}
\[
\max_{\mathcal S}\det(\tilde{\mathbf L}_{\mathcal S}),
\]
\emph{but computes it through the feature factorization $\tilde{\mathbf L}=\mathbf \Phi\mathbf \Phi^\top$.}

\paragraph{Complexity.}
The standard kernel-space Cholesky greedy algorithm explicitly constructs and stores 
$\tilde{\mathbf L} \in \mathbb{R}^{T \times T}$, requiring $\mathcal{O}(T^2 d)$ time for kernel construction and $\mathcal{O}(T^2)$ memory. 
During inference, the $i$-th greedy step updates a length-$T$ basis vector using the previous $i-1$ basis vectors, incurring a cost of $\mathcal{O}(iT)$. 
Summing over $K$ iterations yields a total inference cost of $\mathcal{O}(TK^2)$. Therefore, the overall time complexity is $\mathcal{O}(T^2 d + TK^2),$ with $\mathcal{O}(T^2)$ memory. In contrast, the kernel-free method stores only $\mathbf \Phi\in\mathbb R^{T\times d}$ and $K$ feature-space basis vectors $\{\mathbf c_i\}_{i=1}^K$, which costs $\mathcal{O}(Td+Kd)$ memory. Each iteration requires computing inner products in $d$ dimensions for all $T$ items, leading to $\mathcal{O}(Td)$ work per iteration and $\mathcal{O}(KTd)$ total time. Therefore, when $d\ll T$, the kernel-free formulation is exactly equivalent in target but substantially more efficient.

\section{Proof and Interpretation of Density-aware GD Importance}
\label{appendix:gd_importance}

In this section, we provide a derivation and interpretation of the proposed
Group-DPP (GD) importance score and its density-aware variant used for dynamic
resolution allocation. The key idea is that each selected frame is first
evaluated by its marginal contribution to the determinant of the selected
group. This contribution admits both an algebraic interpretation as a
determinant ratio and a geometric interpretation as residual uniqueness with
respect to the other selected frames. We then incorporate a fixed density prior
to obtain the final score used for budget-aware pruning and token allocation.

\subsection{Definition}

Let $\mathcal S$ be the subset selected by Global Linear DPP. Recall that the
query-conditioned frame feature matrix is
\begin{equation}
\mathbf \Phi
=
\mathrm{diag}(\tilde{\mathbf r})\hat{\mathbf F}
\in \mathbb R^{T\times d},
\end{equation}
where the $t$-th row is denoted by
$\boldsymbol\phi_t^\top\in\mathbb R^d$. For any subset
$\mathcal A$, define the group Gram matrix
\begin{equation}
\mathbf G_{\mathcal A}
=
\mathbf \Phi_{\mathcal A}\mathbf \Phi_{\mathcal A}^{\top}.
\end{equation}


For a selected frame $t\in\mathcal S$, let
\begin{equation}
\mathcal A_t=\mathcal S\setminus\{t\}.
\end{equation}
The pure GD importance is defined as the multiplicative leave-one-out
contribution of frame $t$ to the selected group:

\begin{equation}
\begin{aligned}
\mathcal I_t
&=
\exp\left(
\log\det(\mathbf G_{\mathcal S})
-
\log\det(\mathbf G_{\mathcal A_t})
\right)\\
&=
\frac{
\det(\mathbf G_{\mathcal S})
}{
\det(\mathbf G_{\mathcal A_t})
}.
\end{aligned}
\label{eq:gd_importance_ratio}
\end{equation}

Equivalently, the log GD score is
\begin{equation}
s_t
=
\log \mathcal I_t
=
\log\det(\mathbf G_{\mathcal S})
-
\log\det(\mathbf G_{\mathcal A_t}).
\label{eq:gd_log_score}
\end{equation}

Thus, $s_t$ measures the additive leave-one-out marginal contribution of
frame $t$ to the DPP log-volume objective, while $\mathcal I_t$ measures its
multiplicative contribution to the determinant. In practice, a small diagonal
jitter is added to the Gram matrix when computing log-determinants for
numerical stability.

\subsection{GD Importance as Multiplicative Marginal Contribution}

The determinant $\det(\mathbf G_{\mathcal S})$ measures the squared volume
spanned by the query-conditioned features in $\mathcal S$. Therefore,
Eq.~\eqref{eq:gd_importance_ratio} measures how much this group volume changes
when frame $t$ is included:
\begin{equation}
\det(\mathbf G_{\mathcal S})
=
\det(\mathbf G_{\mathcal A_t})
\cdot
\mathcal I_t.
\end{equation}
A larger $\mathcal I_t$ indicates that frame $t$ contributes more new volume
to the selected group. In contrast, if frame $t$ is redundant with respect to
the other selected frames, its contribution to the determinant becomes small.

\subsection{Residual Uniqueness Interpretation}
\label{appendix:gd_residual_uniqueness}

We next show that the pure GD importance $\mathcal I_t$ is exactly the squared
residual norm of frame $t$ after projecting it onto the span of the other
selected frames.

Assume that $\mathbf G_{\mathcal A_t}$ is non-singular. After reordering the
indices so that frame $t$ is the last element, the Gram matrix of
$\mathcal S=\mathcal A_t\cup\{t\}$ can be written as
\begin{equation}
\mathbf G_{\mathcal S}
=
\begin{bmatrix}
\mathbf G_{\mathcal A_t}
&
\mathbf{\Phi}_{\mathcal A_t}\boldsymbol\phi_t
\\
\boldsymbol\phi_t^{\top}\mathbf{\Phi}_{\mathcal A_t}^{\top}
&
\boldsymbol\phi_t^{\top}\boldsymbol\phi_t
\end{bmatrix}.
\label{eq:gd_block_matrix}
\end{equation}

By the Schur complement formula,
\begin{align}
\det(\mathbf G_{\mathcal S})
=
\det(\mathbf G_{\mathcal A_t})\\
\left(
\boldsymbol\phi_t^{\top}\boldsymbol\phi_t
-
\boldsymbol\phi_t^{\top}
\mathbf{\Phi}_{\mathcal A_t}^{\top}
\mathbf G_{\mathcal A_t}^{-1}
\mathbf{\Phi}_{\mathcal A_t}
\boldsymbol\phi_t
\right).
\end{align}
Dividing both sides by $\det(\mathbf G_{\mathcal A_t})$ gives
\begin{equation}
\mathcal I_t
=
\boldsymbol\phi_t^{\top}
\left(
\mathbf I
-
\mathbf{\Phi}_{\mathcal A_t}^{\top}
\mathbf G_{\mathcal A_t}^{-1}
\mathbf{\Phi}_{\mathcal A_t}
\right)
\boldsymbol\phi_t.
\label{eq:gd_residual_form}
\end{equation}

Define
\begin{equation}
\mathbf P_{\mathcal A_t}
=
\mathbf{\Phi}_{\mathcal A_t}^{\top}
\left(
\mathbf{\Phi}_{\mathcal A_t}\mathbf{\Phi}_{\mathcal A_t}^{\top}
\right)^{-1}
\mathbf{\Phi}_{\mathcal A_t}.
\end{equation}
This is the orthogonal projector onto the span of the features in
$\mathcal A_t$. Therefore,
\begin{equation}
\mathcal I_t
=
\boldsymbol\phi_t^{\top}
(\mathbf I-\mathbf P_{\mathcal A_t})
\boldsymbol\phi_t
=
\left\|
(\mathbf I-\mathbf P_{\mathcal A_t})
\boldsymbol\phi_t
\right\|_2^2.
\label{eq:gd_residual_norm}
\end{equation}
This proves the proposition in the main text. The pure GD importance of frame
$t$ is exactly the amount of information in $\boldsymbol\phi_t$ that cannot be
linearly explained by the other selected frames. Therefore, $\mathcal I_t$
captures residual uniqueness within the query-conditioned feature space.

Since $\boldsymbol\phi_t=\tilde r_t\hat{\mathbf f}_t$ and
$\|\hat{\mathbf f}_t\|_2=1$, we also have
\begin{equation}
0
\le
\mathcal I_t
\le
\|\boldsymbol\phi_t\|_2^2
=
\tilde r_t^2.
\label{eq:gd_upper_bound}
\end{equation}
Thus, the pure GD score is upper-bounded by the query relevance of the frame.
A frame can receive a high pure GD importance only when it is both
query-relevant and not redundant with respect to the rest of the selected
group.

\subsection{Density-aware GD Score}
Although pure GD importance captures residual uniqueness, it may still
overemphasize frames that are geometrically distinctive but have relatively
weak query-conditioned feature magnitude. To stabilize the score used for
dynamic resolution, we incorporate a fixed density prior:
\begin{equation}
\rho_t
=
\frac{
\|\boldsymbol\phi_t\|_2^2
}{
\frac{1}{|\mathcal S|}
\sum_{j\in\mathcal S}
\|\boldsymbol\phi_j\|_2^2
}.
\label{eq:gd_density_prior}
\end{equation}
The final density-aware GD score is
\begin{equation}
\tilde{\mathcal I}_t
=
\mathcal I_t \rho_t.
\label{eq:density_aware_gd_score_appendix}
\end{equation}

Because $\boldsymbol\phi_t=\tilde r_t\hat{\mathbf f}_t$ and
$\|\hat{\mathbf f}_t\|_2=1$, the density prior is proportional to the squared
query relevance:
\begin{equation}
\|\boldsymbol\phi_t\|_2^2 = \tilde r_t^2.
\end{equation}
Therefore, $\tilde{\mathcal I}_t$ combines two factors:
\begin{equation}
\tilde{\mathcal I}_t
=
\underbrace{
\left\|
(\mathbf I-\mathbf P_{\mathcal A_t})
\boldsymbol\phi_t
\right\|_2^2
}_{\text{residual uniqueness}}
\cdot
\underbrace{
\rho_t
}_{\text{query-density prior}}.
\end{equation}
This score preserves the diversity-aware residual contribution of GD while
further favoring frames with strong query-conditioned magnitude. In the rest
of the allocation procedure, we use $\tilde{\mathcal I}_t$ rather than the
pure GD score $\mathcal I_t$.

\subsection{Why Density-aware GD Encourages Relevant Diversity}

The residual form in Eq.~\eqref{eq:gd_residual_norm} directly explains why GD
importance discourages redundancy. If frame $t$ is nearly explained by the
other selected frames, then
\begin{equation}
(\mathbf I-\mathbf P_{\mathcal A_t})\boldsymbol\phi_t
\approx
\mathbf 0,
\end{equation}
and therefore $\mathcal I_t$ is small. In contrast, if frame $t$ contains
information that is not covered by the other selected frames, its residual
norm is large and $\mathcal I_t$ increases.

The density prior further modulates this residual uniqueness by the magnitude
of the query-conditioned feature. Therefore, a frame receives a large
density-aware GD score only when it is both non-redundant with respect to the
selected group and strongly aligned with the query. This avoids assigning
excessive visual tokens to low-density outliers whose residual uniqueness is
large only because they are visually unusual.

Moreover, if $\mathcal A\subseteq\mathcal B$, then the span of
$\mathcal A$ is contained in the span of $\mathcal B$. Hence projecting onto
$\mathcal B$ can only reduce the residual:
\begin{equation}
\left\|
(\mathbf I-\mathbf P_{\mathcal B})
\boldsymbol\phi_t
\right\|_2^2
\le
\left\|
(\mathbf I-\mathbf P_{\mathcal A})
\boldsymbol\phi_t
\right\|_2^2.
\end{equation}
This gives a diminishing-residual interpretation: as more similar frames are
included, the marginal uniqueness of an additional frame decreases.

\section{Implementation and Evaluation Details}
\label{appendix:implementation}
\subsection{Evaluation Setup}
All evaluations are conducted with the official \texttt{lmms-eval} pipeline. 
We use the default prompt templates, answer extraction rules, and scoring scripts provided by each benchmark. 
Unless otherwise specified, all methods use the same decoding configuration for a given backbone and benchmark.

During evaluation, no additional training, subtitles, audio inputs, or external tools are used. 
For each video-query pair, different frame selection methods are applied before feeding the selected visual inputs into the same target MLLM. For methods that require frame-query similarity, we use the same LongCLIP features to avoid differences caused by external encoders.

For closed-source models, we do not access model internals or modify the inference process. 
LDDR is used only as an external preprocessing step: it selects query-relevant and diverse frames, assigns different resolutions according to their GD importance scores, and sends the resulting multi-resolution frames with the text query to the model API.

\paragraph{Inference setting.}
All experiments are conducted under a unified inference setting using the official \texttt{lmms-eval} evaluation pipeline. We use greedy decoding for all open-source MLLMs, with temperature set to 0, top-$p$ disabled, beam size set to 1, and a maximum generation length of 128 tokens. No subtitles, audio inputs, external tools, or additional training data are used during evaluation. For video preprocessing, we decode each video into frames at 1 FPS and apply the corresponding frame selection strategy before feeding the selected visual inputs into the target MLLM.

For all methods, the total visual budget is normalized by the number of full-resolution frame equivalents. Following the Qwen-VL visual-token setting, one full-resolution frame corresponds to 1024 visual tokens, and each visual token corresponds to a $14 \times 14$ image patch. Therefore, a frame budget of $F$ corresponds to a total budget of $F \times 1024$ visual tokens. For fixed-resolution methods, each selected frame is assigned 1024 tokens. For dynamic-resolution methods, the per-frame token budget is constrained to $w_{\min}=256$ and $w_{\max}=1024$, while keeping the total visual-token budget identical across methods.

All experiments are evaluated on NVIDIA RTX A6000 GPUs. We use LongCLIP as the default text-image encoder for frame pre-encoding across our method and all frame-selection baselines, unless otherwise specified. The evaluated MLLM backbones include Qwen2.5-VL-7B, Qwen3-VL-8B, LLaVA-OneVision-1.5-8B, InternVL3.5-4B, and InternVL3.5-8B.

\subsection{Candidate Frame Pool and Feature Extraction}

For each benchmark, we first sample candidate frames from each video at 1 FPS. 
All frame selection methods are then applied to this shared candidate frame pool to select the final visual inputs for the target MLLM. 
This ensures that different methods are compared under the same initial temporal coverage and that performance differences mainly come from the selection strategy rather than the candidate frame construction.

\subsection{Benchmark Prompt Templates}
\label{appendix:prompts}

To improve the reproducibility of our evaluation protocol, we provide examples of the prompt templates used in our experiments. 
These prompts follow the official benchmark or \texttt{lmms-eval} formats whenever available, and are used consistently across different frame selection methods and MLLM backbones. 
\begin{tcolorbox}[title=Video-MME, colback=gray!15, colframe=black!70, boxrule=0.5pt]
\ttfamily
<video>\\
Select the best answer to the following multiple-choice
question based on the video and the subtitles. Respond with
only the letter (A, B, C, or D) of the correct option.\\
\{question\}\\
\{option\_1\}\\
\{option\_2\}\\
\{option\_3\}\\
\{option\_4\}\\
Answer with the option's letter from the given choices directly.
\end{tcolorbox}

\begin{tcolorbox}[title=LongVideoBench, colback=gray!15, colframe=black!70, boxrule=0.5pt]
\ttfamily
<video>\\
Select the best answer to the following multiple-choice
question based on the video and the subtitles. Respond with
only the letter (A, B, C, D or E) of the correct option.\\
\{question\}\\
A. \{option0\}\\
B. \{option1\}\\
C. \{option2\}\\
D. \{option3\}\\
E. \{option4\_if\_exists\}\\
Answer with the option's letter from the given choices directly.
\end{tcolorbox}

\begin{tcolorbox}[title=LVBench, colback=gray!15, colframe=black!70, boxrule=0.5pt]
\ttfamily
<video>\\
\{question\}\\
Answer the question with the option letter
\end{tcolorbox}

\begin{tcolorbox}[title=MLVU, colback=gray!15, colframe=black!70, boxrule=0.5pt]
\ttfamily
<video>\\
\{question\}\\
Only give the best option.\\
Best option:
\end{tcolorbox}
\end{document}